\documentclass[conference]{IEEEtran}
\IEEEoverridecommandlockouts
\usepackage{cite}
\usepackage{amsmath,amssymb,amsfonts}
\usepackage{algorithmic}
\usepackage{graphicx}
\usepackage{textcomp}
\usepackage{xcolor}
\def\BibTeX{{\rm B\kern-.05em{\sc i\kern-.025em b}\kern-.08em
    T\kern-.1667em\lower.7ex\hbox{E}\kern-.125emX}}

\usepackage{algorithm}
\usepackage{algorithmic}
\usepackage{tabularx}
\usepackage{multirow}
\usepackage{multicol}
\usepackage{subfigure}

\newcommand{\etal}{\textit{et al.}}
\newcommand{\ie}{\textit{i.e.}}
\begin{document}

\title{XPipe: Efficient Pipeline Model Parallelism for Multi-GPU DNN Training\thanks{This work was done during the visit of the first author to UCLA.}
}

\author{\IEEEauthorblockN{Lei Guan\textsuperscript{\rm 1}, Wotao Yin\textsuperscript{\rm 2}, Dongsheng Li\textsuperscript{\rm 1}, Xicheng Lu\textsuperscript{\rm 1}}
\IEEEauthorblockA{\textit{\textsuperscript{\rm 1}College of Computer, National University of Defense Technology} \\
\textit{\textsuperscript{\rm 2}Department of Mathematics, University of California, Los Angeles}\\
guanleics@gmail.com, wotaoyin@math.ucla.edu, \{dsli, xclu\}@nudt.edu.cn}
}

\maketitle

\begin{abstract}
We propose XPipe, an efficient asynchronous pipeline model parallelism approach for multi-GPU DNN training. XPipe is designed to use multiple GPUs to concurrently and continuously train different parts of a DNN model. To improve GPU utilization and achieve high throughput, it splits a mini-batch into a set of micro-batches. It allows the overlapping of the pipelines of multiple micro-batches, including those belonging to different mini-batches. Most importantly, the novel weight prediction strategy adopted by XPipe enables it to effectively address the weight inconsistency and staleness issues incurred by the asynchronous pipeline parallelism. As a result, XPipe incorporates the advantages of both synchronous and asynchronous pipeline model parallelism approaches. Concretely, it can achieve very comparable (even slightly better) model accuracy as its synchronous counterpart while obtaining higher throughput than it. Experimental results show that XPipe outperforms other state-of-the-art synchronous and asynchronous model parallelism approaches.
\end{abstract}


\section{Introduction}
Deep Neural Networks (DNNs) have been recognized as one of the most effective tools for many machine learning tasks including image and video analysis~\cite{szegedy2017inception,karpathy2014large}, language translation~\cite{kalchbrenner2014convolutional,wu2016google}, and speech recognition~\cite{afouras2018deep,lee2009unsupervised}. Training a DNN model, however, often takes hours, days, and even weeks. The long training time of DNN models is mainly because the training always involves a huge amount of data and a large number of parameters (also known as weights) \cite{you2018imagenet,shazeer2017outrageously}.

In the past few years, one needs to scale up DNNs as they are used to simultaneously recognize more subjects or objects~\cite{taigman2014deepface}, which requires the DNNs to have many more layers and weights. Such increases inevitably create a higher demand for the memory of the training devices and the training throughput. Sometimes, breaking a DNN model into pieces and training them with multiple GPUs come to be the only choice for training a neural network with a vast amount of parameters.
%, for example, the ImageNet~\cite{deng2009imagenet}, JFT \cite{hinton2015distilling}, and OpenImages \cite{krasin2017openimages} datasets

Data parallelism is the most commonly used approach to utilize multiple GPU devices to accelerate DNN training. For data parallelism, each GPU holds a full copy of the DNN weights and is assigned with a subset of training data. Weight update happens only when the gradients on all GPUs are aggregated. Another orthogonal approach is model parallelism~\cite{ben2018demystifying}, where the DNN structure is divided into subsets of layers and assigned to different GPUs. The naive model parallelism strategy divides the DNN into a set of submodels (each including one or more consecutive layers) and assigns each submodel to a GPU device~\cite{lee2014model}. Each GPU only computes and transmits the activation to the next GPU in the forward direction, unless it owns the last layer, and computes and transmits gradients to the previous GPU in the backward direction unless it keeps the first layer. The inter-GPU communication overhead in model parallelism can be much less than that in data parallelism. However, the naive approach always works serially. That is, in each feedforward-backpropagation round, after a GPU completes its forward step, it waits until all its subsequent GPUs finish their forward and backward steps before it starts its backward step. This leads to the sequential activation of the GPUs and causes serious under-utilization of the multi-GPU system.

To this end, we propose XPipe, an efficient asynchronous pipeline model parallelism approach. This work is motivated by the state-of-the-art synchronous pipeline approach GPipe \cite{huang2019gpipe}, as well as asynchronous pipeline, approaches PipeDream~\cite{narayanan2019pipedream} and SpecTrain\cite{chen2018efficient}, which will be detailedly reviewed in the next section. XPipe inherits the pipeline structure of PipeDream and SpecTrain but uses a micro-batch as the basic processing unit and adopts a more efficient strategy to address the weight inconsistency and staleness issues incurred by the asynchronous pipeline parallelism. Besides, adopting fine-grained micro-batch also makes XPipe easily scale up the mini-batch size. Even though XPipe introduces micro-batches into the pipeline training as GPipe, it allows the cross-training of these micro-batches from different mini-batches, giving rise to better GPU utilization and higher throughput than GPipe. In summary, XPipe incorporates the advantages of both synchronous and asynchronous pipeline model parallelism approaches. It provides high throughput, scales up mini-batch size easily, and achieves very comparable (even slightly better) model accuracy as its synchronous counterpart (\ie, GPipe).

We evaluated XPipe using three popular Convolutional Neural Network (CNN) models on two different image datasets. The experimental results are detailedly reported, which demonstrate the effectiveness of our proposal. In comparison to PipeDream and SpecTrain, XPipe effectively alleviates the accuracy degradation and achieves very comparable (even slightly better) model accuracy as GPipe. At the same time, XPipe can obtain consistently higher throughput than GPipe, regardless of the number of mini-batch partitions. For example, when training Inception-V3 on Tiny ImageNet, XPipe provides an average of 20.0\% (up to 31.9\%) and 88.1\% (up to 150.8\%) throughput improvement over GPipe on 2-GPU and 4-GPU systems, relatively.

\section{Related Work}

A great many works have explored model parallelism to accelerate DNN training using multiple GPUs~\cite{dean2012large,huo2018training,huo2018decoupled,pal2019optimizing}. Besides, pipelining has also been heavily studied and widely applied to speed up neural network training~\cite{petrowski1993performance,kamruzzaman2013load,chen2012pipelined,pittman2018exploring}. Combining the advantages of both model parallelism and pipelining, pipeline model parallelism has been recently proposed to efficiently train DNNs in a model-parallel manner~\cite{huang2019gpipe,narayanan2019pipedream}. According to the way the weights are updated, existing pipeline model parallelism approaches can be roughly classified into two categories: synchronous pipeline model parallelism and asynchronous pipeline model parallelism.

\textit{Synchronous pipeline model parallelism}. The state-of-the-art synchronous approach is GPipe~\cite{huang2019gpipe}, which was proposed to address the low GPU-utilization problem of the naive model parallelism strategy and overcome the memory limitation for scaling up DNNs. The noteworthy feature of GPipe is that it first splits a mini-batch into a set of smaller micro-batches. Therefore, the training has a finer data unit; each mini-batch is trained equally by training a set of micro-batches. %Each micro-batch gets $\frac{N}{T}$ training examples; and
Introducing micro-batches into the pipeline training makes GPipe pretty good at scaling up the mini-batch size. More importantly, GPipe trains each set of micro-batches in a pipelined manner, which, to some extent, allows the concurrent training of multiple GPUs.
In this mean, GPU utilization is significantly improved compared to the naive model parallelism strategy. Meanwhile, GPipe belongs to the synchronous-parallel approach and thus can train DNNs without degrading their model accuracy.
However, since these micro-batches from the same mini-batch must flow through all the GPUs sequentially, GPipe cannot always keep all GPUs busy with training the model in parallel, so it still suffers from load imbalance.

\textit{Asynchronous pipeline model parallelism}. Asynchronous model-parallel (AMP) training~\cite{gaunt2017ampnet} was proposed to overcome the low device-utilization problem in the naive model parallelism as well. AMP training allows asynchronous (thus faster) weight update as long as enough gradients are accumulated. However, AMP faces serious weight inconsistency and staleness issues due to the cross-training of multiple mini-batches.
Besides that, another asynchronous pipeline parallel approach called PipeDream~\cite{narayanan2019pipedream} was recently proposed. Similar to AMP training, PipeDream introduces multiple workers' concurrent processing by simultaneously training multiple mini-batches in the pipeline. To address the weight inconsistency issue incurred by the cross-training of multiple mini-batches, PipeDream keeps a copy of the weights for each mini-batch active in the pipeline. However, keeping the weights wastes GPU memory, especially when a DNN model has a massive amount of model parameters. On the other hand, PipeDream suffers from a staleness problem because it uses different weights in the whole feedforward-backpropagation round \cite{chen2018efficient}. The staleness issue slows down the convergence and degrades the model accuracy as well. To simultaneously alleviate the inconsistency and staleness issues in the asynchronous pipeline model parallelism, Chen \etal~\cite{chen2018efficient} proposed SpecTrain. It adopts the same pipeline structure as PipeDream, enables cross-training multiple mini-batches, and achieves high GPU utilization. Instead of storing the weights for each active mini-batch in the pipeline, SpecTrain addresses weight inconsistency and staleness issues. Based on the observation that the smoothed gradients used in Momentum SGD \cite{qian1999momentum} reflect the trend of weight updates, in both forward and backward passes, SpecTrain uses the smoothed gradients time the weights version difference to predict the future weights. However, as shown in the experiments later, SpecTrain can still not completely solve the inconsistency and staleness issues and often incurs accuracy drop.

\section{Method}
%This section describes the main method of XPipe. The symbols used in this paper are summarized in Table \ref{tab:symbol}.

%\begin{table}[!htp]
%\centering
%\caption{Symbols.}
%\label{tab:symbol}
%\begin{tabular}{cc}
%\hline
%Symbol & Explanation           \\
%\hline
%$rank$   & GPU index  \\
%$size$   & total \# of GPUs  \\
%$N$  & mini-batch size \\
%$T$  & \#micro-batches of a mini-batch ($N\%T$=$0$)  \\
%$n$  & total \#micro-batches  \\
%$W_t$  & model weights at pipeline time $t$   \\
%$\hat{W_t}$  & predicted weights corresponding to $W_t$   \\	
%$v_t$  & smoothed gradients   \\	
%$m_t$  & average of squared gradients\\
%$s$  & predicted-actual weight difference \\
%$fwd\_idx$   &  index of micro-batch in forward pass  \\
%$bwd\_idx$   &   index of micro-batch in backward pass  \\
%\hline
%\bottomrule[0.55pt]
%\end{tabular}
%\end{table}

%%
%\begin{figure*}[!htbp]
%\centerline{\includegraphics[width=0.80\textwidth, height=3.0cm]{figure/gpipe}}
%\caption{Illustration of GPipe on the 4-GPU platform ($T=4$). The blue dashed lines indicate the weight update. }
%\label{gpipe}
%\end{figure*}
%%

%\begin{figure}[!htbp]
%\includegraphics[width=0.42\textwidth, height=3.8cm]{fly}
%\caption{Illustration of the \emph{draining phase} of XPipe.}
%\label{xpipe-drain}
%\end{figure}

%\begin{figure}
%\centering
%\includegraphics{fly}
%\caption{A sample black and white graphic (.pdf format).}
%\label{fig:fly}
%\end{figure}

\begin{figure*}[!htbp]
	\centering
	%\subfigure[real\_sim]{\includegraphics[width=0.3\textwidth, height=4.3cm]{news20-acc.eps}\label{real_sim_breakdown}}
	\subfigure[XPipe with $T=2$]{\includegraphics[width=0.80\textwidth, height=4.5cm]{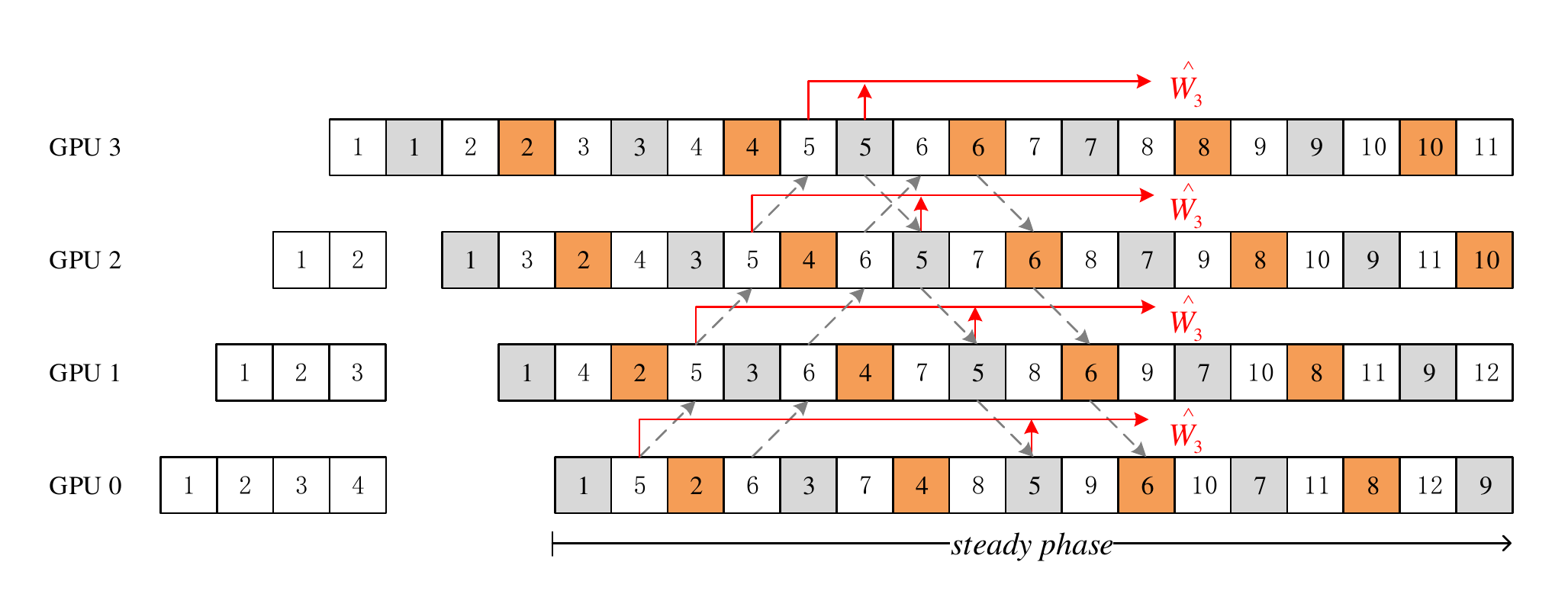}\label{xpipe-p2}}
	\subfigure[XPipe with $T=4$]{\includegraphics[width=0.80\textwidth, height=4.5cm]{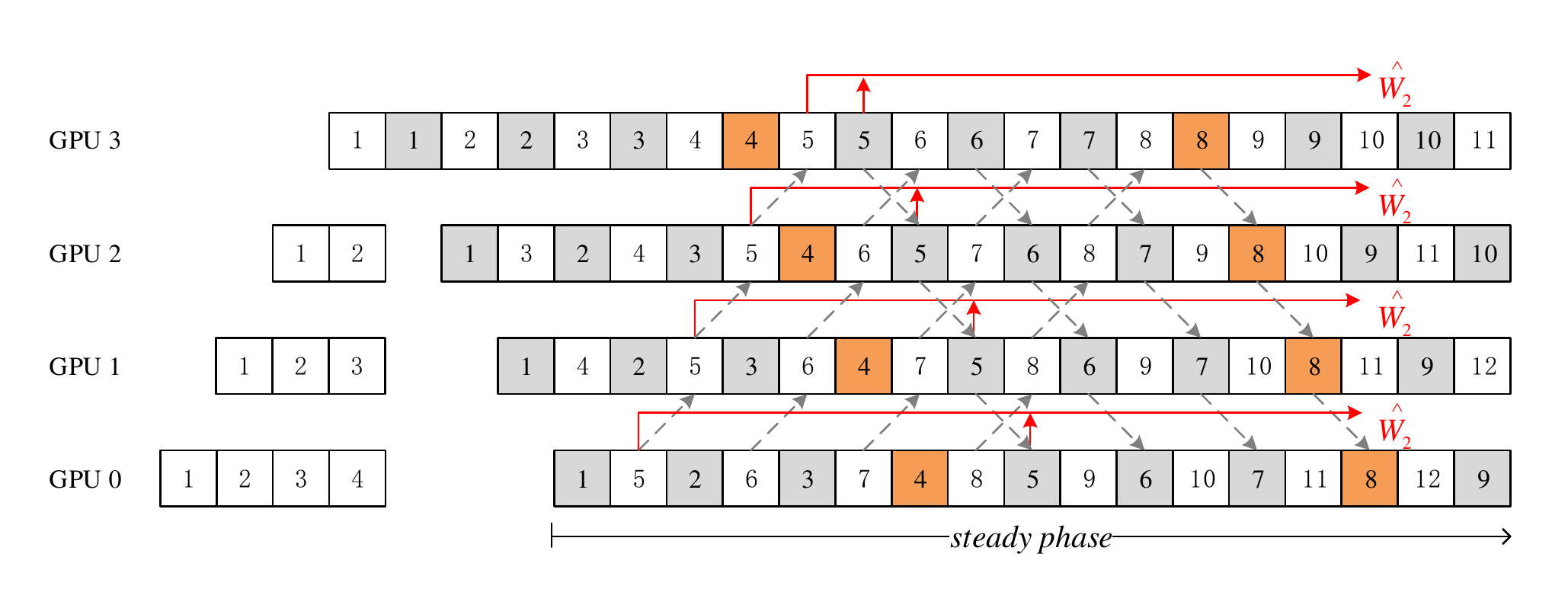}\label{xpipe-p4}}
	\caption{Illustration of XPipe workflow on the 4-GPU system. Top: XPipe workflow with $T=2$; bottom: XPipe workflow with $T=4$. %The number $i$ inside each box refer to the computation of the $i$-th micro-batch. The white boxes denote the forward pass; grey boxes indicate the backward pass; orange boxes refer to the backward pass of the last micro-batch,  at the end of which weights are updated. The grey dashed lines in Figure \ref{xpipe-p2} depict the round trip of processing the 3-\emph{rd} mini-batch (\ie, micro-batches 5 and 6); the grey dashed lines in Figure \ref{xpipe-p4} depict the round trip of processing the 2-\emph{nd} mini-batch (\ie, micro-batches 5, 6, 7 and 8). The red solid lines shows the weight prediction performed by the \emph{bellwether}.
	}
	\label{xpipe}
\end{figure*}

\subsection{Workflow}\label{sec:xpipe-workflow}
%XPipe is built based on the same pipeline structure as PipeDream and SpecTrain but with different training workflow.
In XPipe, each mini-batch of size $N$ is split into $T$ smaller micro-batches. Thus a micro-batch of size $N/T$ becomes the basic data processing unit throughout the pipeline training. Figures \ref{xpipe-p2} and \ref{xpipe-p4} illustrate the workflow of XPipe on the 4-GPU system with $T=2$ and $T=4$, respectively. The number $i$ inside each box refer to the forward or backward pass of the $i$th micro-batch. The white boxes denote the forward passes; grey boxes indicate the backward passes; orange boxes refer to the backward passes of the $T$th micro-batches, at the end of which weights are updated. The grey dashed lines with arrows in Figure \ref{xpipe-p2} depict the round trip of processing the third mini-batch (\ie, micro-batches 5 and 6); the grey dashed lines with arrows in Figure \ref{xpipe-p4} depict the round trip of processing the second mini-batch (\ie, micro-batches 5, 6, 7 and 8). In the workflow of XPipe, each mini-batch is trained equally through the training of $T$ micro-batches.
For example, in Figure \ref{xpipe-p2}, micro-batches 1 and 2 correspond to mini-batch 1, and so on. Similarly, for XPipe with $T=4$ (as shown in the bottom figure of Figure \ref{xpipe-p4}), micro-batches 1 to 4 correspond to mini-batch 1, and so on. The red arrowed lines in Figures \ref{xpipe-p2} and \ref{xpipe-p4} depict the weight prediction, which will be detailedly described later.

One noteworthy feature of XPipe is that the $T$ micro-batches corresponding to the same mini-batch should share the same weights in their forward and backward passes. Weight update does not instantly happen when a micro-batch completes its backward pass.  Instead, when performing the backward pass, the gradients are consistently accumulated across micro-batches and applied to update model parameters only when the $T$th micro-batch completes its backward pass (as shown by the orange boxes in Figure \ref{xpipe}). %This leads to a result that the weight update frequency decrease with the setting of $T$. For example, for the case when $T=1$,

Beyond that, as depicted in Figures \ref{xpipe-p2} and \ref{xpipe-p4}, XPipe intersects the execution order of micro-batches belonging to different mini-batches. That is, XPipe allows the cross-training of mini-batches, which is quite different from GPipe. In this way, all GPUs can continuously and concurrently train their submodels after the \emph{steady phase} starts, giving rise to pretty high GPU utilization. Unfortunately, the cross-training of micro-batches results in weight inconsistency and staleness issues. For example, in Figure \ref{xpipe-p2}, GPU 0 uses the initial weights to perform the forward pass of the fifth micro-batch. However, when GPU 0 is ready to run the backward pass of it, the weights on GPU 0 have been updated twice, \ie, after the backward passes of micro-batches 2 and 4.
Moreover, as shown in Figures \ref{xpipe-p2} and \ref{xpipe-p4}, for each micro-batch, the number of weight updates happened between the forward pass and corresponding backward pass varies with the index of GPU. The GPUs with smaller index tend to use staler weights to perform the forward and backward passes. The staleness issue further slows down the convergence and hurts the model quality at the same time.

\subsection{Weight Prediction}
This section proposes an efficient weight prediction strategy to simultaneously address the weight inconsistency and staleness issues arising in the asynchronous pipeline training. Instead of using the smoothed gradients, XPipe performs weight prediction based on Adam~\cite{kingma2014adam} updates, where a running average of the first and second moment of the gradients are used.

For $T$ micro-batches corresponding to a mini-batch, we refer to the micro-batch with the minimum index as a \emph{bellwether}. Each mini-batch is allocated with a \emph{bellwether} being in charge of making weight prediction. For instance, the \emph{bellwether} of the third mini-batch in Figure \ref{xpipe-p2} is micro-batch 5 and the \emph{bellwether} of the second mini-batch in Figure \ref{xpipe-p4} is micro-batch 5 as well. The noteworthy feature of a \emph{bellwether} is that it always comes in first to perform both forward and backward passes among the $T$ micro-batches.%In XPipe, we let the
We use the weights version difference $s$ to measure the number of weight updates between the current pipeline unit and the pipeline unit at which the $T$-th micro-batch on GPU 0 completes its training round trip. The version difference $s$ should always be calculated first when the \emph{bellwether} is ready to perform weight prediction.

For forward pass, the \emph{bellwether} calculates the version difference via
\begin{equation}
s=round(\frac{size+ T-rank/2-2}{T}),
\label{step}
\end{equation}
where $size$ refers to the amount of GPUs and $rank$ is the index of each GPU.

At the backward pass, the version difference turns to
\begin{equation}
s=round(\frac{T + \left\lfloor rank/2 \right\rfloor - 1}{T}).
\label{step1}
\end{equation}

For both forward and backward passes, the \emph{bellwether} of the $t$-th mini-batch uses following formula to predict the corresponding future weights:
%V_{t-1}
\begin{equation}
\hat{W}_{t}= W_t - s\cdot lr\cdot\Delta W_t,
\label{weight_update}
\end{equation}
where $lr$ is the learning rate and $\Delta W_t= \frac{\overline{v}_{t}}{\sqrt{\overline{m}_t + \epsilon}}$ with
\begin{equation}
\left\{\begin{array}{ll}
g_t = \nabla (W_t), \\
v_t = \gamma \cdot v_{t-1} + (1-\gamma)\cdot g_t, \\
\overline{v}_{t}= \frac{v_t}{1-\gamma}, \\
{m}_{t}= \lambda\cdot m_{t-1} + (1-\lambda) \cdot g_t^2,\\
\overline{m}_{t}= \frac{m_{t}}{1-\lambda}.\\
\end{array}\right.
\label{formular_group}
\end{equation}
In (\ref{formular_group}), $g_t$ refers to the gradients of stochastic objective corresponding to the $t$-th mini-batch; $v_t$ is the biased first-moment estimate; $m_t$ is the biased second raw moment estimate; $\overline{v}_{t}$ is the bias-corrected first-moment estimate; $\overline{m}_{t}$ is the bias-corrected second raw moment estimates; $g_t^2$ refers to element-wise square with $g_t^2=g_t\odot g_t$; $\epsilon$, $\gamma$ and $\lambda$ are constant values.

%In Equation (\ref{weight_update}), the version difference $s$ indicates the amount of weight updates between current pipeline unit and the pipeline unit at which the mini-batch completes its train round. The version difference $s$ should be calculated first by the \emph{bellwether} applies Equation (\ref{weight_update}) to do weight prediction.

%For weight prediction at both forward and backward passes,

Figures \ref{xpipe-p2} and \ref{xpipe-p4} illustrate the main idea of weight prediction by the \emph{bellwether} on 4-GPU system with $T=2$ and $T=4$ respectively. The red arrowed lines stand for the weight prediction performed by the \emph{bellwether}. All of them start from the pipeline unit where the \emph{bellwether}s start their forward passes and point to the pipeline unit at which their corresponding mini-batches on GPU 0 finish the whole train round. In Figures \ref{xpipe-p2} and \ref{xpipe-p4}, $\hat{W_t}$ denotes the predicted weights corresponding to the $t$th mini-batch.
%whole round trip of the mini-batch ends.
On each GPU, when $T$ micro-batches (\ie, a mini-batch) are ready to perform the forward pass or the backward pass in sequence, the \emph{bellwether} will first calculate the version difference $s$; then weight prediction is performed using the current weights $W_t$ and version difference $s$ to generate the future weights $\hat{W_t}$ through~(\ref{weight_update}). Following that, the other ($T-1$) micro-batches will directly apply $\hat{W_t}$ to perform their forward or backward passes.

In the following, we illustrate the weight prediction procedure of XPipe using the pipeline training procedure with $T=4$ on the 4-GPU system. As shown in Figure \ref{xpipe-p4}, on GPU 0, when the second mini-batch (\ie, micro-batches 5, 6, 7 and 8) is ready to perform the forward pass, micro-batch 5 should first use formula~(\ref{step}) to calculate the version difference $s$, and then apply formula~(\ref{weight_update}) to calculate the future weights for the second mini-batch (\ie, $\hat{W_2}$). After that, micro-batches 6, 7, and 8 directly make use of $\hat{W_2}$ to perform their forward passes. To avoid repeatedly making weight predictions, the $\hat{W_2}$ generated by the \emph{bellwether} (\ie, micro-batch 5) should be temporarily cached and then directly utilized by the following ($T-1$) micro-batches.
% otherwise, the same weight prediction procedure will have to be repeated multiple times.
%Meanwhile, it is worth noting that an ordinary micro-batch may occasionally play the role of \emph{bellwether} when another mini-batch updates the weights in the middle.
%update of bursts into the execution sequence of the forward passes of these micro-batches.
%As shown in Figure \ref{xpipe-p4}, on GPU 2, weights update for the 1-\emph{st} mini-batch happens when micro-batch 4 completes its backward pass. Then micro-batches 6, 7 and 8 cannot use the cached predicted weights generated by micro-batch 5. In this case, micro-batch 6 acts as a \emph{bellwether} and performs the weight prediction using formula~(\ref{weight_update}). %Then the version  difference $s$ generated by Equation (\ref{step}) should subtract 1 in this case.
%After that, the subsequent micro-batches (micro-batches 7 and 8) will inherit the latest predicted weights (generated by micro-batch 6) to perform their forward passes.

Likewise, at the backward pass, a \emph{bellwether} again takes charge of predicting future weights. As shown in Figure \ref{xpipe-p4}, when micro-batch 5 is ready to perform the backward pass, it will first use formula (\ref{step1}) to calculate the version difference $s$, and then apply~(\ref{weight_update}) to predict the future weights $\hat{W_2}$. As with the prediction in the forward pass, the generated weights via prediction are first cached and then reused by the subsequent ($T-1$) micro-batches for their backward passes to avoid repetitive weight predictions.

\section{Experimental Results}
\subsection{Implementation Details}
We implemented XPipe using PyTorch~\cite{paszke2017automatic} of version 1.2.0.
% The code of XPipe will be released on Github.
In the implementation of XPipe, each GPU was allocated with one process. Each process was in charge of managing the local memory, data transfer between the host and GPUs, gradients calculation, weight update, as well as communicating with other processes. PyTorch provides a package called \emph{torch.distributed} for process-to-process message passing. In XPipe, each process used the MPI communication backend to realize inter-GPU communication. Non-block communication primitives (e.g., $\textit{isend}$ and $\textit{irecv}$) were used to overlap inter-GPU communication and GPU computation.

\subsection{Model Partition}
The premise of pipeline model parallelism is to partition a DNN model into a set of submodels. A few prior works are concentrating on efficient partitioning \cite{mirhoseini2017device,narayanan2019pipedream,huang2019gpipe}.
Designing an efficient partitioning algorithm is beyond the focus of this paper. In the experiments, we always partition all the DNN layers across GPUs with a roughly equal number of layers to balance their training time while letting the latter GPUs have a slightly greater number of layers to achieve time/memory balance across GPUs.

\subsection{Experiment Setup}
We conducted all the experiments on a 4-GPU system equipped with 4 GeForce RTX2080X Nvidia GPUs. The host CPU is an Intel i9-9940X (@3.30 GHz).

Three popular CNN models were chosen as the benchmark networks in our experiments: VGG-16 \cite{simonyan2014very}, ResNet-101 \cite{he2016deep} and Inception-V3 \cite{Szegedy_2016_CVPR}. Two image datasets were used in the experiments. The first dataset is CIFAR-10~\cite{krizhevsky2009learning} which includes 60000 32$\times$32 images in total, 50000 images for training and 10000 images for validation. The second dataset is Tiny ImageNet~\cite{yao2015tiny}, which is categorized into 200 classes each having 500 training images and 50 validation images. Standard data augmentation schemes, including flipping, padding and random crop, were used in both of these two datasets. To be concrete, the CIFAR-10 images were normalized using mean = [0.4914, 0.4822, 0.4465] and std = [0.2023, 0.1994, 0.2010]. For Tiny ImageNet, each $64\times 64\times 3$ image was first scaled up to $224\times 224\times 3$.  Following that, the images were loaded into a range of [0, 1] and then normalized using mean=[0.485, 0.456, 0.406] and std=[0.229, 0.224, 0.225].

%In order to verify this, we also construct XPipe's weight prediction strategy using the smoothed gradients as in DualPipe, which we call XPipe*. We found that, XPipe* converges much faster than DualPipe despite that both of their weight prediction strategies are constructed using the smoothed gradients.

In the experiments, we compared XPipe with following state-of-the-art pipeline model parallelism approaches: PipeDream (with weight stashing)~\cite{narayanan2019pipedream}, SpecTrain~\cite{chen2018efficient} and GPipe~\cite{huang2019gpipe}. The following three measures were taken to ensure a fair comparison:
1. As with XPipe, we implemented PipeDream, SpecTrain, and GPipe using the PyTorch framework;
2. Before the pipeline training starts, all the evaluated methods adopted the same model partitioning approach to split all the DNN layers across GPUs;
3. Each evaluated approach took advantage of the same strategy (\ie, automatically reperformed the forward pass during the backward pass~\cite{huang2019gpipe}) for better memory utilization.
In all the experiments, we let the first GPU read the training data while letting the last GPU read the corresponding ground-truth labels. The seed was fixed with 1 for shuffling the data. For XPipe, we empirically set $\gamma=0.9$, $\lambda=0.999$ and $\epsilon=1e-8$. Meanwhile, the elements of both $v_t$ and $m_t$ were initialized using $1e-4$ times randomly generated numerical values ranging from 0 to 1.

\subsection{Results and Discussions}
\textbf{Comparison of XPipe, PipeDream, and SpecTrain} In this section, we compared XPipe with PipeDream and SpecTrain in terms of convergence and model accuracy. Since GPipe without mini-batch partitioning automatically reduces to the naive pipeline approach, we trained GPipe with $T=1$ to simulate the behavior of the naive approach and saw the learning results of it as the baseline. We also trained XPipe with $T=1$ to isolate the effects of model partition. We selected VGG-16 and Inception-V3 as the benchmark network and used 4 GPUs to train them on CIFAR-10 for 90 epochs. The learning rate was initialized as 1e-2, and divided by 10 every 30 epochs.
%The total images are divided into 100 classes each containing 600 images. We use 50000 images  %The total images are divided into 100 classes each containing 600 images. We use 50000 images for training and 10000 images for validation.
%As with CIFAR-10 dataset, same data augmentation are applied here.
%For CIFAR-10, we trained ResNet-101 and Inception-V3 for 120 epochs with t \footnote{https://github.com/kuangliu/pytorch-cifar} \footnote{https://github.com/weiaicunzai/pytorch-cifar100}
We trained the model using the Momentum SGD with the momentum factor $\gamma$ was set to 0.9, and weight decay was 5e-4. The mini-batch size for all the evaluated methods was 128.

\begin{figure}[htbp]
	\centering
	%\subfigure[ResNet-101 on CIFAR-10]{\includegraphics[width=0.24\textwidth, height=3.8cm]{figure/sgd_resnet101_4_loss}\label{sgd_resnet101_loss}}
	%\subfigure[ResNet-101 on CIFAR-10]{\includegraphics[width=0.24\textwidth, height=3.8cm]{figure/sgd_resnet101_4_top1}\label{sgd_resnet101_top1}}
	\subfigure{\includegraphics[width=0.23\textwidth, height=3.4cm]{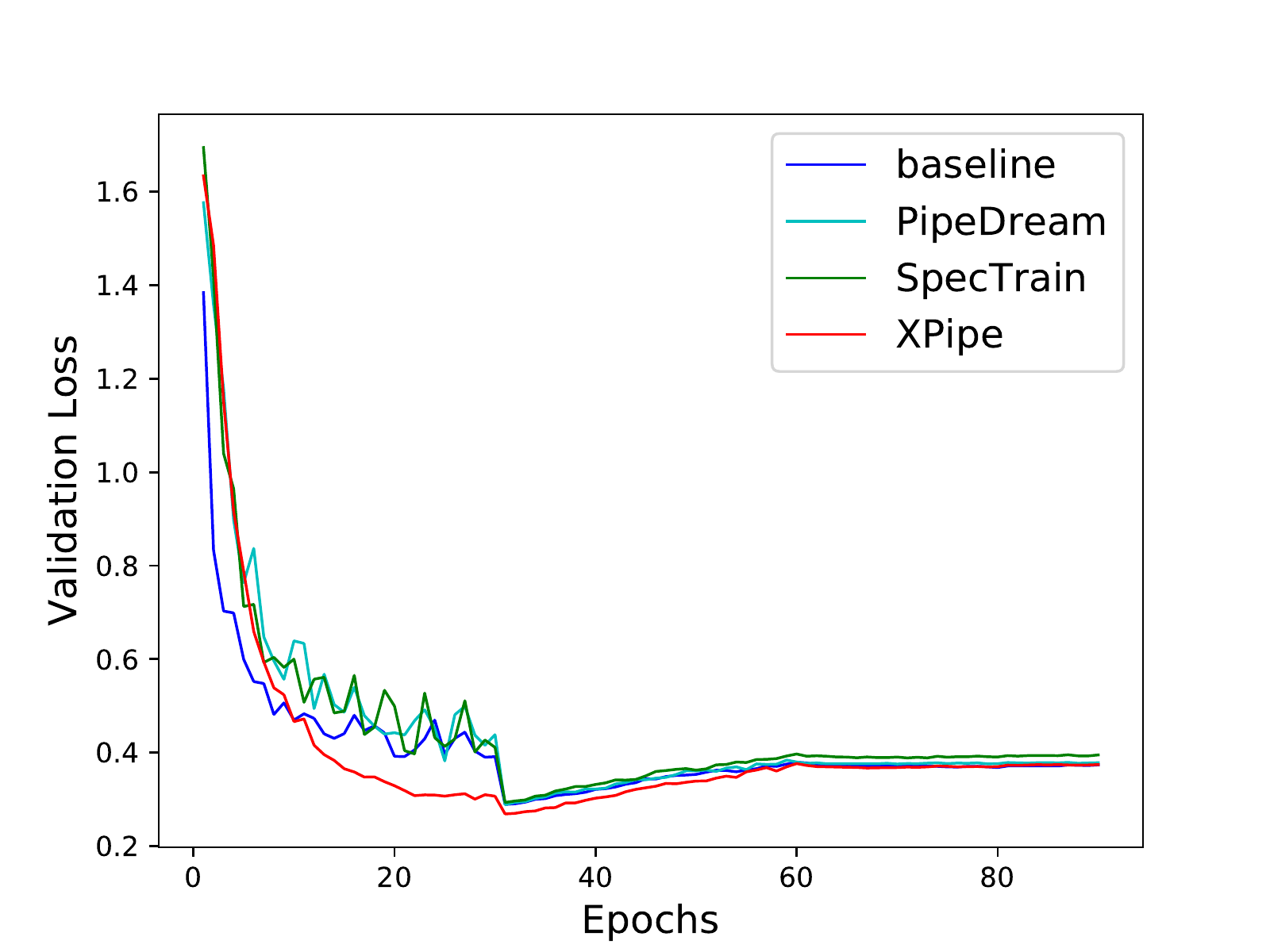}\label{sgd_vgg16_loss}}
	%\subfigure[Inception-V3]{\includegraphics[width=0.23\textwidth, height=3.4cm]{figure/async_sgd_inceptionv3_4_loss}\label{sgd_vgg16_loss}}
	\subfigure{\includegraphics[width=0.23\textwidth, height=3.4cm]{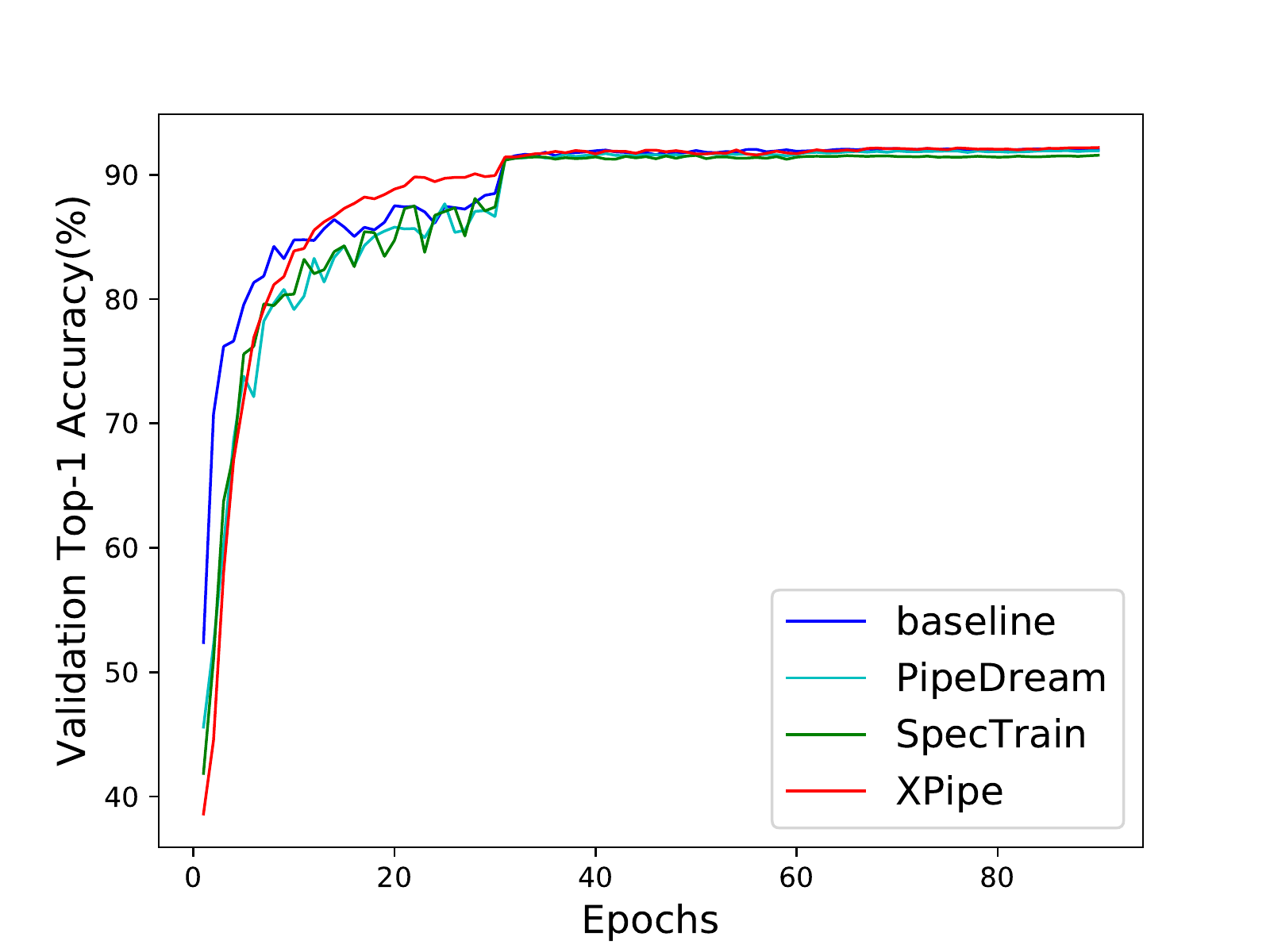}\label{sgd_vgg16_top1}}
	%\subfigure[Inception-V3]{\includegraphics[width=0.23\textwidth, height=3.4cm]{figure/async_sgd_inceptionv3_4_top1}\label{sgd_vgg16_top1}}
	%\subfigure[VGG-16]{\includegraphics[width=0.23\textwidth, height=3.4cm]{figure/async_sgd_vgg16_4_top1}\label{sgd_inceptionv3_top1}}
	%\subfigure[Inception-V3]{\includegraphics[width=0.22\textwidth, height=3.4cm]{figure/async_sgd_inceptionv3_4_top1}\label{sgd_vgg16_top1}}
	\caption{Learning curves when training VGG-16 on CIFAR-10. Left: validation loss versus epochs; right: validation accuracy (top-1, in \%) versus epochs. }
	\label{fig:sync-vgg16}
\end{figure}

\begin{figure}[htbp]
	\centering
	%\subfigure[ResNet-101 on CIFAR-10]{\includegraphics[width=0.24\textwidth, height=3.8cm]{figure/sgd_resnet101_4_loss}\label{sgd_resnet101_loss}}
	%\subfigure[ResNet-101 on CIFAR-10]{\includegraphics[width=0.24\textwidth, height=3.8cm]{figure/sgd_resnet101_4_top1}\label{sgd_resnet101_top1}}
	%\subfigure[VGG-16]{\includegraphics[width=0.23\textwidth, height=3.4cm]{figure/async_sgd_vgg16_4_loss}\label{sgd_inceptionv3_loss}}
	\subfigure{\includegraphics[width=0.23\textwidth, height=3.4cm]{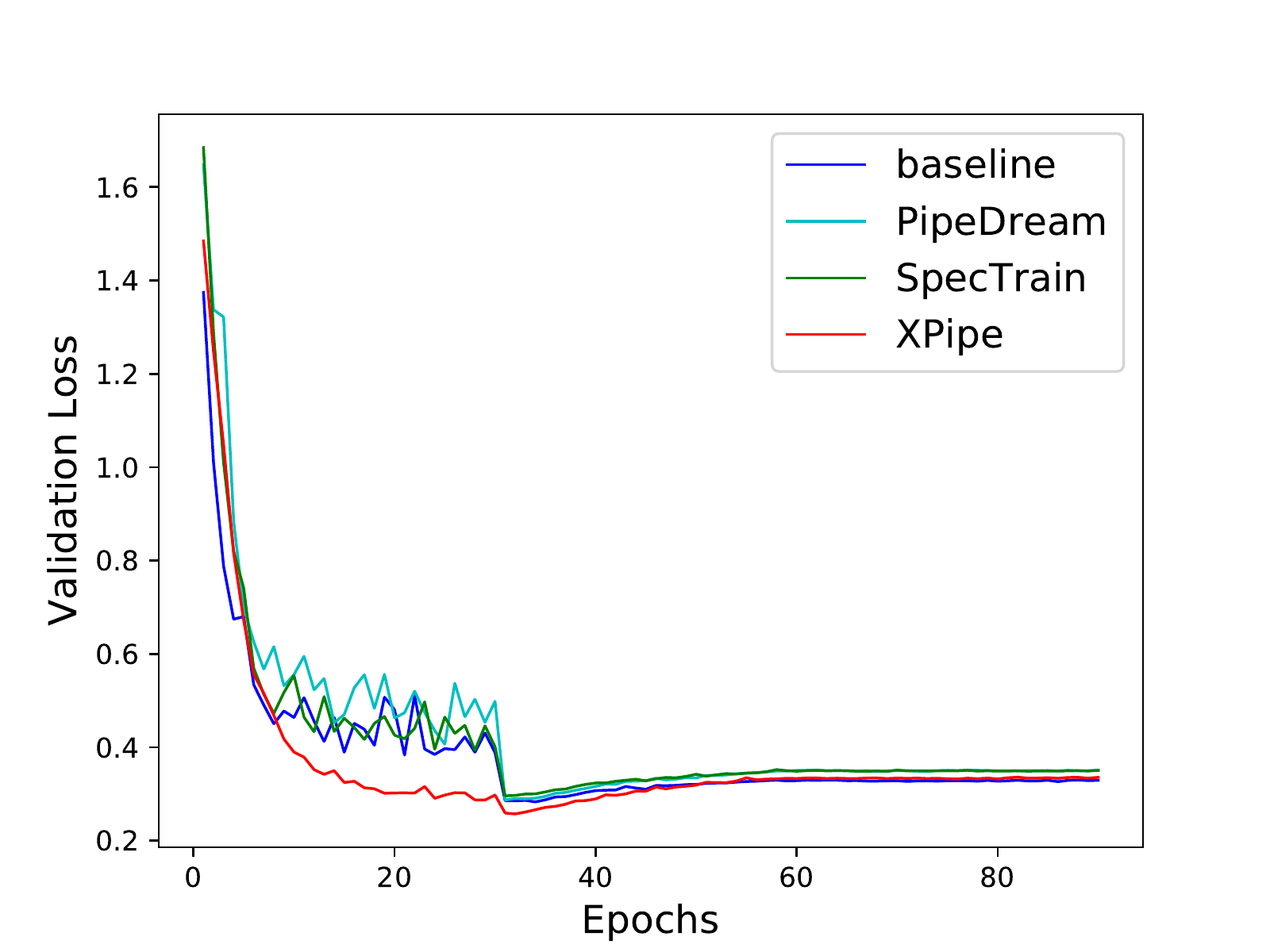}\label{sgd_inceptionv3_loss}}
	%\subfigure[VGG-16]{\includegraphics[width=0.23\textwidth, height=3.4cm]{figure/async_sgd_vgg16_4_top1}\label{sgd_inceptionv3_top1}}
	\subfigure{\includegraphics[width=0.23\textwidth, height=3.4cm]{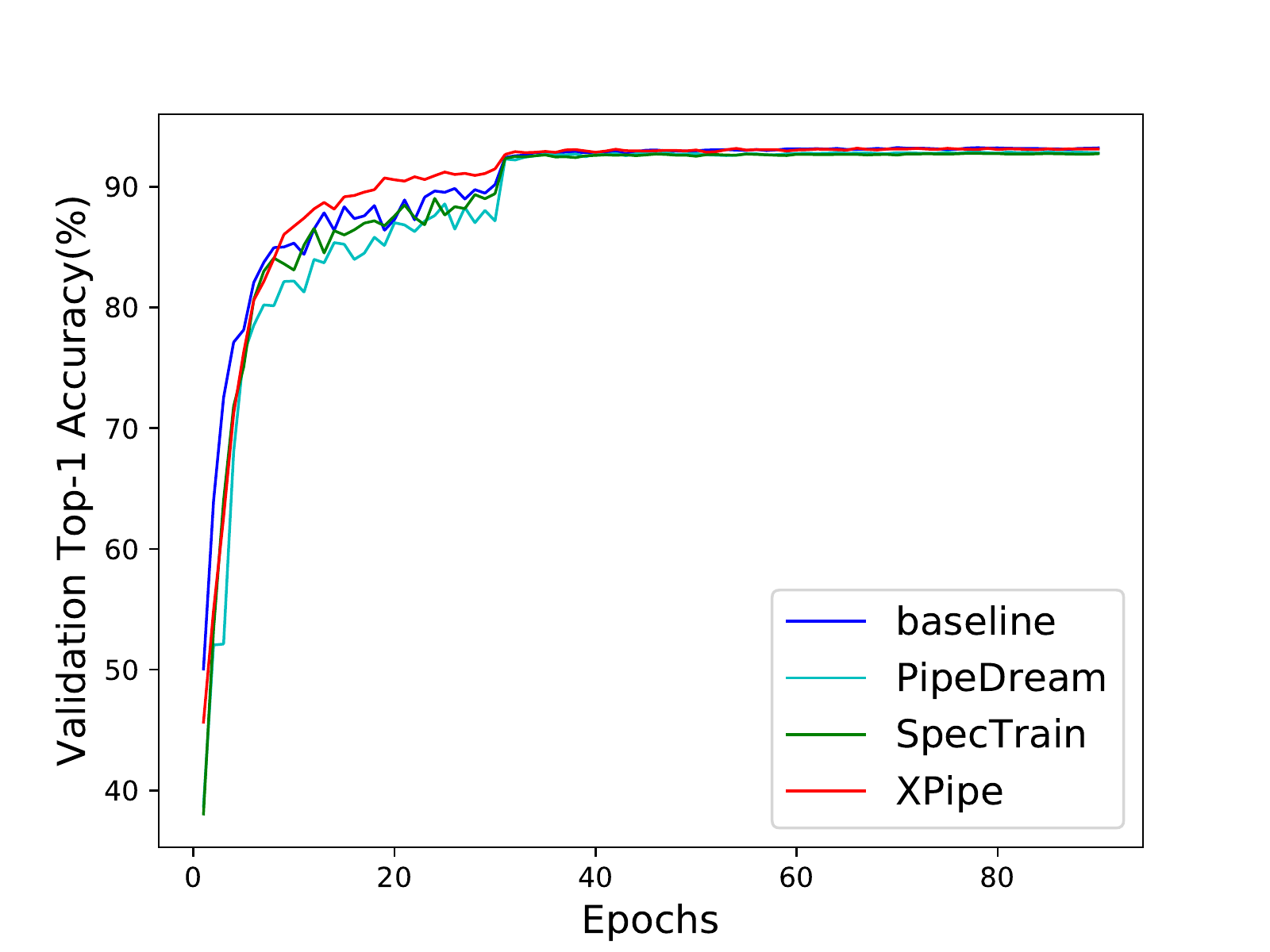}\label{sgd_inceptionv3_top1}}
	%\subfigure[VGG-16]{\includegraphics[width=0.23\textwidth, height=3.4cm]{figure/async_sgd_vgg16_4_top1}\label{sgd_inceptionv3_top1}}
	%\subfigure[Inception-V3]{\includegraphics[width=0.22\textwidth, height=3.4cm]{figure/async_sgd_inceptionv3_4_top1}\label{sgd_vgg16_top1}}
	\caption{Learning curves when training Inception-V3 on CIFAR-10. Left: validation loss versus epochs; right: validation accuracy (top-1, in \%) versus epochs.}
	\label{fig:sync-inceptionv3}
\end{figure}

Figure \ref{fig:sync-vgg16} depicts the learning curves for training VGG-16; Figure \ref{fig:sync-inceptionv3} shows the learning curves for training Inception-V3. Table \ref{tab:sync} summarizes the obtained minimal validation loss and maximum validation top-1 accuracy. XPipe converges very fast, and its learning curves converge even slightly better than that of the baseline. Besides, the experimental results show that XPipe can always obtain the least validation loss value and very comparable validation top-1 accuracy as the baseline. On average, XPipe achieves 0.015\% top-1 validation accuracy improvement over the baseline. In contrast, PipeDream and SpecTrain incur an average of 0.265\% and 0.51\% top-1 accuracy drop, respectively. Note that when running on the 4-GPU system, XPipe with $T=1$ generates the same version differences as SpecTrain to make the weight prediction in both the forward and backward passes. Still, it converges faster and obtains better model accuracy than SpecTrain. The experiment results verify that the Adam-based weight prediction provides a more effective solution for weight prediction.

\begin{table}[htbp]
	\caption{Results on CIFAR-10. Top: results for training VGG-16; bottom: results for training Inception-V3. The values inside the parentheses denote the relative variation of validation top-1 accuracy compared to the results generated by the baseline. The best results are highlighted in boldface.}
	\begin{center}
		\begin{tabular}{c||ll}
			\hline
			\multirow{2}{*}{Approach}
			& Min. Val. & Max. Val. \\
			& Loss & Top-1 Accuracy \\
			\hline
			baseline & 0.289 & 92.10\% ($\sim$) \\
			PipeDream &0.289 & 91.93\% (-0.17\%)\\
			SpecTrain &0.293 & 91.56\% (-0.54\%)    \\
			XPipe &\textbf{0.269} & \textbf{92.18\%} (+0.08\%)  \\
			\hline
			\hline
			baseline & 0.283 & \textbf{93.26\%} ($\sim$) \\
			PipeDream &0.287 & 92.90\% (-0.36\%)\\
			SpecTrain &0.296 & 92.78\% (-0.48\%)    \\
			XPipe &\textbf{0.257} & 93.21\% (-0.05\%)  \\
			\hline
			%\hline
			%baseline & 0.283 & \textbf{93.26\%} ($\sim$) \\
			%PipeDream &0.287 & 92.90\% (-0.36\%)\\
			%SpecTrain &0.296 & 92.78\% (-0.48\%)    \\
			%XPipe &\textbf{0.257} & 93.21\% (-0.05\%)  \\
			%\hline
		\end{tabular}
		\label{tab:sync}
	\end{center}
\end{table}

\begin{figure*}[htbp]
	\centering
	%\subfigure[ResNet-101 on CIFAR-10]{\includegraphics[width=0.24\textwidth, height=3.8cm]{figure/sgd_resnet101_4_loss}\label{sgd_resnet101_loss}}
	%\subfigure[ResNet-101 on CIFAR-10]{\includegraphics[width=0.24\textwidth, height=3.8cm]{figure/sgd_resnet101_4_top1}\label{sgd_resnet101_top1}}
	%\subfigure[T=1]{\includegraphics[width=0.33\textwidth, height=3.4cm]{figure/sync_sgd_inceptionv3_4_loss_t1}\label{sgd_inceptionv3_loss_t1}}
	%\subfigure[T=2]{\includegraphics[width=0.33\textwidth, height=3.4cm]{figure/sync_sgd_inceptionv3_4_loss_t1}\label{sgd_inceptionv3_loss_t2}}
	%\subfigure[T=4]{\includegraphics[width=0.33\textwidth, height=3.4cm]{figure/sync_sgd_inceptionv3_4_loss_t4}\label{sgd_inceptionv3_loss_t4}}
	\subfigure{\includegraphics[width=0.32\textwidth, height=3.8cm]{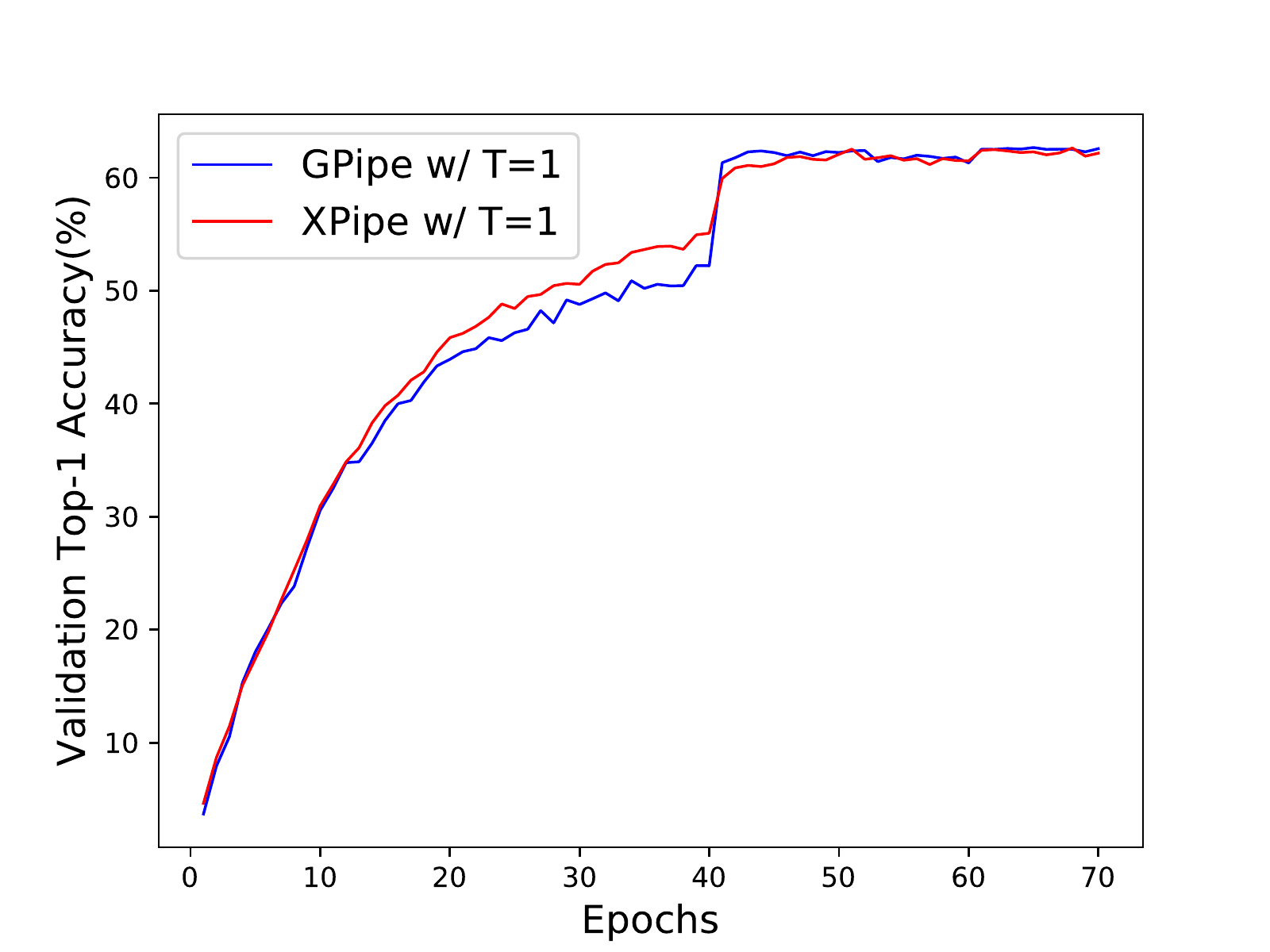}\label{sgd_inceptionv3_top1_t1}}
	\subfigure{\includegraphics[width=0.32\textwidth, height=3.8cm]{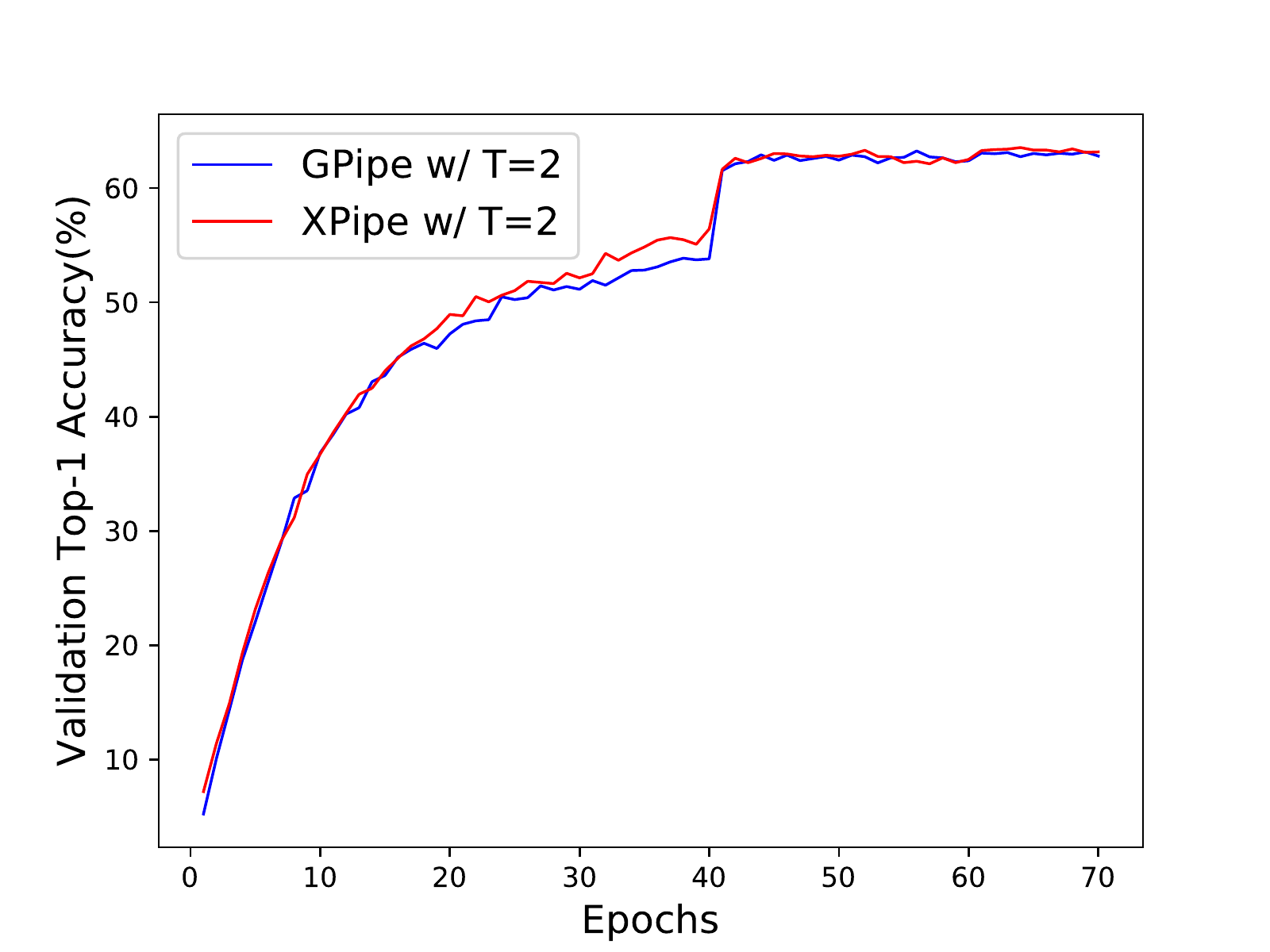}\label{sgd_inceptionv3_top1_t2}}
	\subfigure{\includegraphics[width=0.32\textwidth, height=3.8cm]{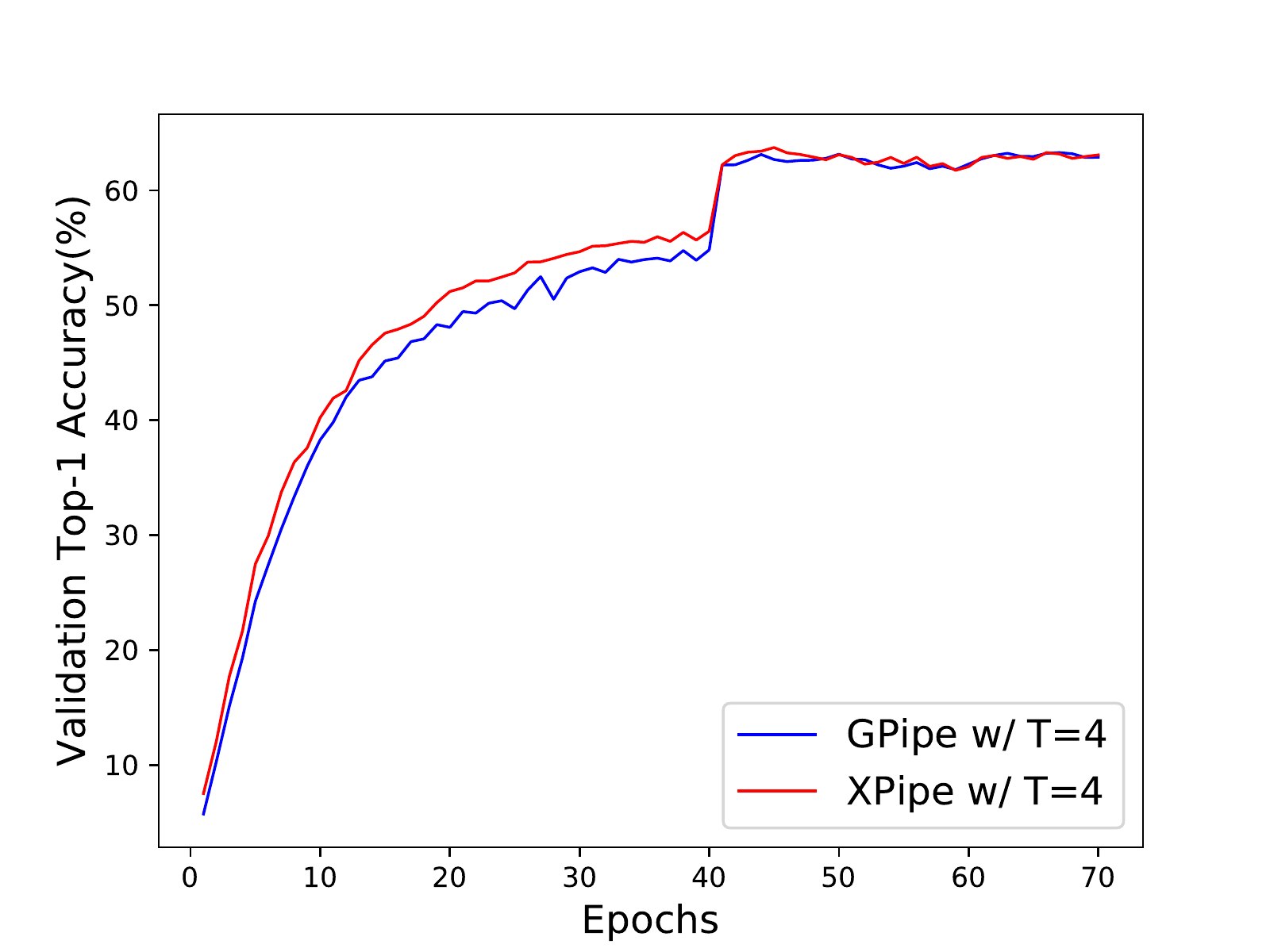}\label{sgd_inceptionv3_top1_t4}}
	\caption{Validation accuracy (top-1, in \%) versus epochs for training Inception-V3 on Tiny ImageNet, with $T=1$, $T=2$, and $T=4$.}
	\label{fig:inceptionv3}
\end{figure*}

\begin{figure*}[htbp]
	\centering
	%\subfigure[ResNet-101 on CIFAR-10]{\includegraphics[width=0.24\textwidth, height=3.8cm]{figure/sgd_resnet101_4_loss}\label{sgd_resnet101_loss}}
	%\subfigure[ResNet-101 on CIFAR-10]{\includegraphics[width=0.24\textwidth, height=3.8cm]{figure/sgd_resnet101_4_top1}\label{sgd_resnet101_top1}}
	%\subfigure[T=1]{\includegraphics[width=0.33\textwidth, height=3.4cm]{figure/sync_sgd_resnet101_4_loss_t2}\label{sgd_resnet101_loss_t1}}
	%\subfigure[T=2]{\includegraphics[width=0.33\textwidth, height=3.4cm]{figure/sync_sgd_resnet101_4_loss_t2}\label{sgd_resnet101_loss_t2}}
	%\subfigure[T=4]{\includegraphics[width=0.33\textwidth, height=3.4cm]{figure/sync_sgd_resnet101_4_loss_t4}\label{sgd_resnet101_loss_t4}}
	\subfigure{\includegraphics[width=0.32\textwidth, height=3.8cm]{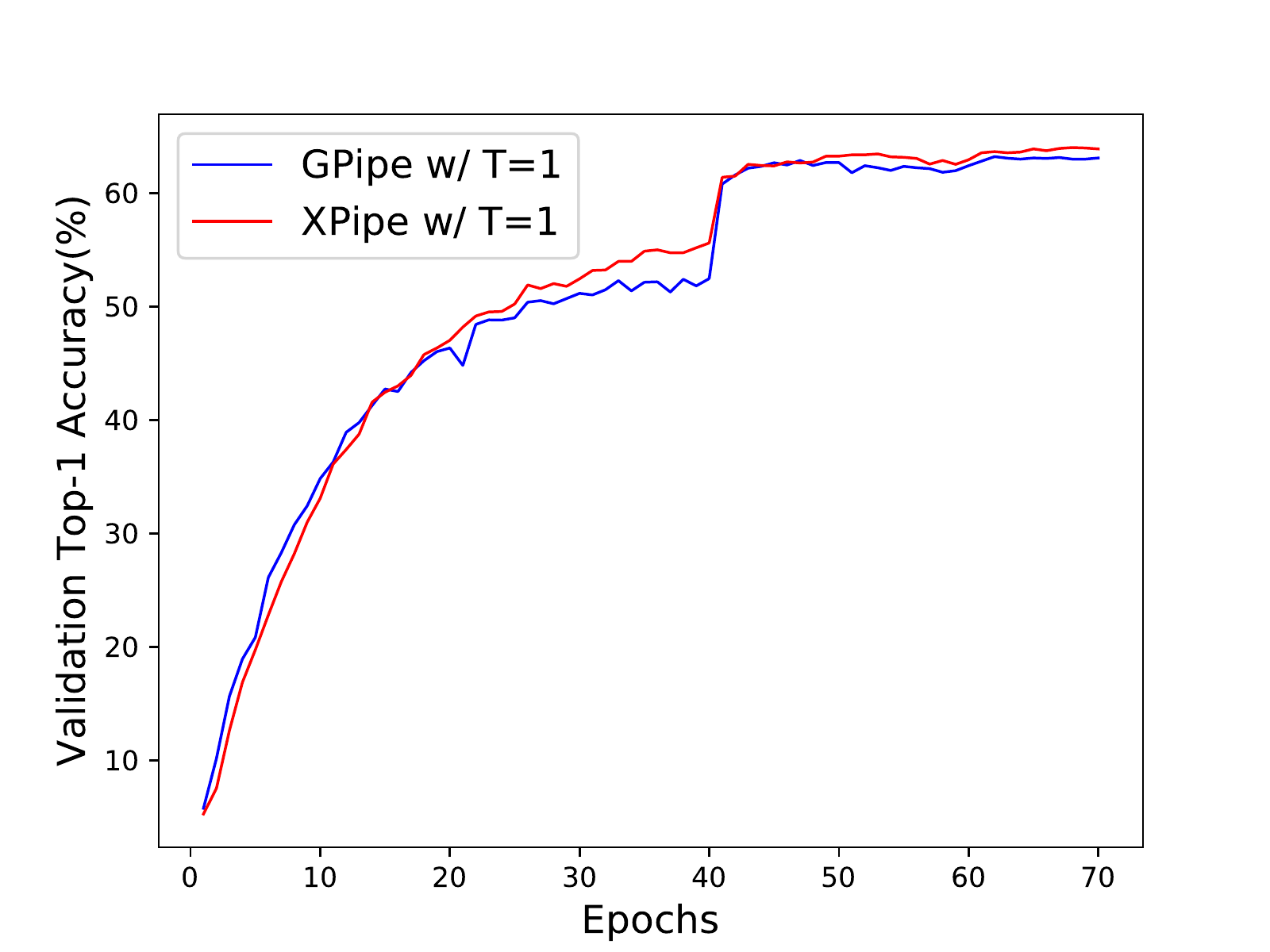}\label{sgd_resnet101_top1_t1}}
	\subfigure{\includegraphics[width=0.32\textwidth, height=3.8cm]{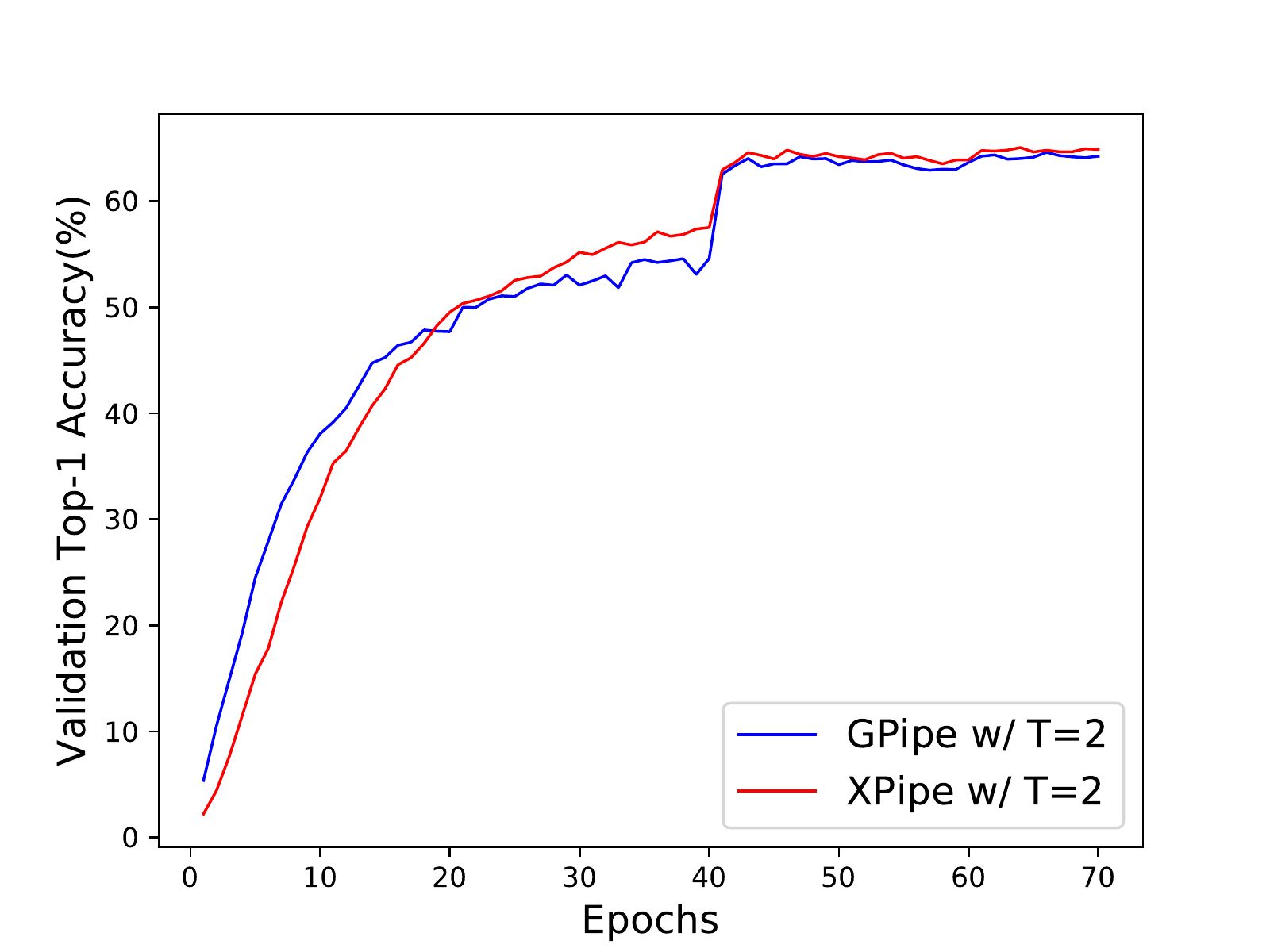}\label{sgd_resnet101_top1_t2}}
	\subfigure{\includegraphics[width=0.32\textwidth, height=3.8cm]{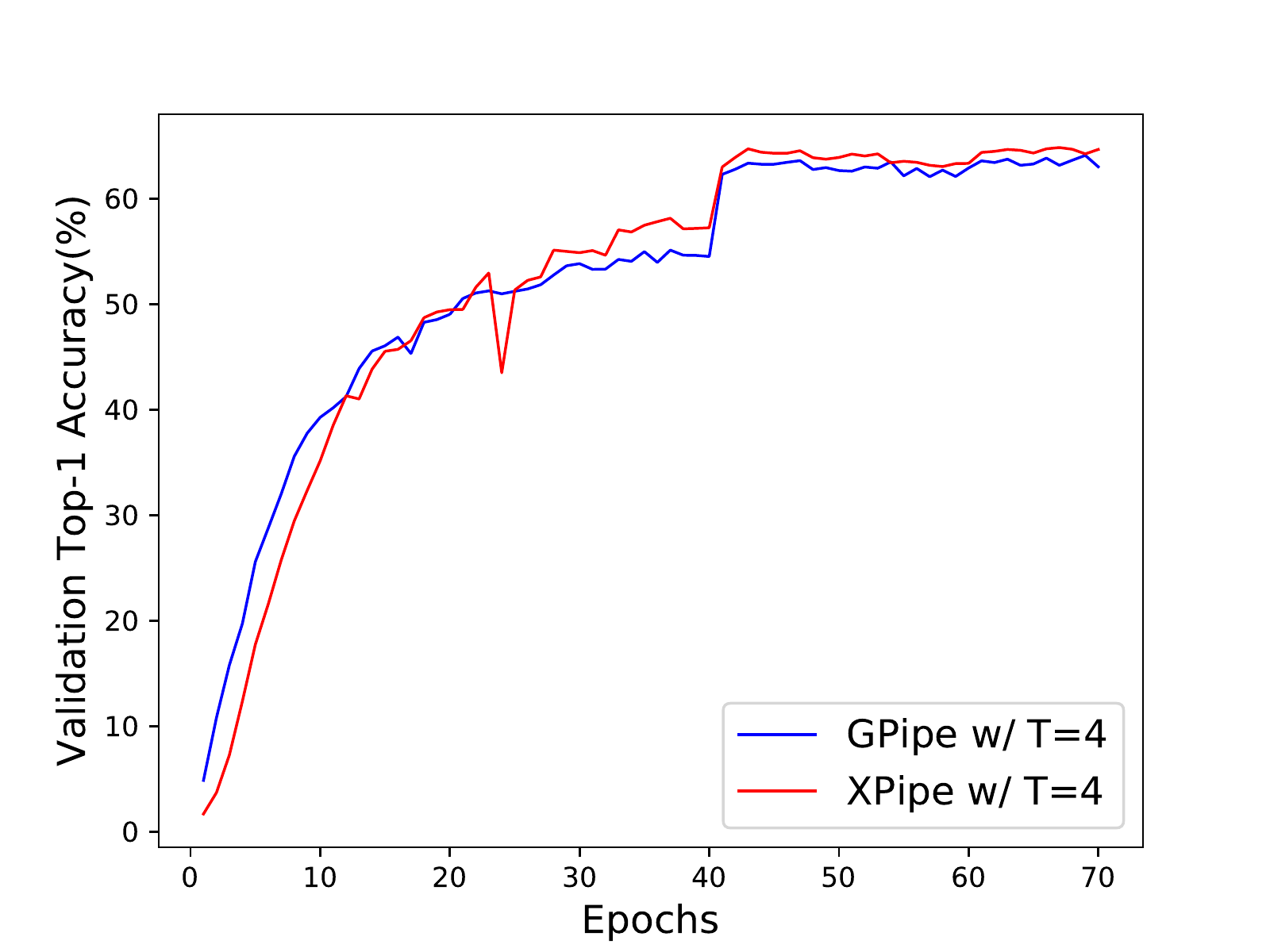}\label{sgd_resnet101_top1_t4}}
	\caption{Validation accuracy (top-1, in \%) versus epochs for training ResNet-101 on Tiny ImageNet, with $T=1$, $T=2$, and $T=4$.}
	\label{fig:resnet101}
\end{figure*}

\noindent\textbf{Comparison with GPipe} In this section, we compared the convergence and model accuracy of XPipe with that of GPipe. We selected Inception-V3 and ResNet-101 as the benchmark network and used 4 GPUs to train them on Tiny ImageNet for 70 epochs.  %The total images are divided into 100 classes each containing 600 images. We use 50000 images for training and 10000 images for validation.
%As with CIFAR-10 dataset, same data augmentation are applied here.
%For CIFAR-10, we trained ResNet-101 and Inception-V3 for 120 epochs with t
We compared XPipe and GPipe by running them with $T=1$, $T=2$ and $T=4$ respectively. In all the experiments, we always set XPipe and GPipe with the same hyper-parameters to compare their performance. The mini-batch size for both of them was fixed with 100. The learning rate was initialized as 1e-2 and divided by 10 at the 40th and 60th epoch.   We trained the models using the Momentum SGD with the momentum factor $\gamma$ was set to 0.9 and weight decay was 5e-4.
%For CIFAR-100, we trained ResNet101 and Inception-V3 for 200 epochs with the learning rate was initialized as 0.1 and divided by 5 at 60th, 120th, 160th epochs.
%For XPipe, as suggested in the literature~\cite{kingma2014adam}, we set $\gamma=0.9$, $\lambda=0.999$ and $\epsilon=1e-8$. Meanwhile, both $v_t$ and $m_t$ were initialized using $1e-4$ times randomly generated values whose value are between 0 and 1. Note that for XPipe and GPipe, we let the number of partitions be equal to the amount of GPUs (\ie, $T=size=4$) in order to isolate the effects of pipeline parallelism.

Figures \ref{fig:inceptionv3} and \ref{fig:resnet101} depict learning curves about the validation accuracy (top-1, in \%) over epochs when three different mini-batch partitions (\ie, $T=1$, $T=2$, and $T=4$) are used. The obtained minimum validation loss and maximum validation top-1 accuracy are reported in Table \ref{tab:tinyimagenet}. For any setting of $T$, XPipe converges very fast and its learning curves on both Inception-V3 and ResNet-101 match well (even converge faster) with that of GPipe. This results again demonstrate the learning-effectiveness of XPipe. Table \ref{tab:tinyimagenet} shows that XPipe almost always achieves smaller validation loss value and higher validation top-1 accuracy than GPipe. On average, XPipe is able to obtain 0.26\% and 0.67\% top-1 validation accuracy improvement over GPipe for training Inception-V3 and ResNet-101 respectively.
%XPipe converges very fast and its learning curves matches well with that of GPipe. In addition, the experimental results show that XPipe is able to abtain .

\begin{table}[htbp]
	\caption{Results on Tiny ImageNet. Top: results for Inception-V3; bottom: results for ResNet-101. The values inside the parentheses denote the relative variation of validation top-1 accuracy compared to the results produced by GPipe. The best results are highlighted in boldface.}
	\begin{center}
		\begin{tabular}{cc||ll}
			\hline
			\multirow{2}{*}{Partition} & \multirow{2}{*}{Method}
			& Min. Val. & Max. Val. \\
			& & Loss & Top-1 Acc. \\
			\hline
			\multirow{2}{*}{$T=1$} & GPipe & \textbf{1.543} & \textbf{62.66\%} ($\sim$) \\
			& XPipe & 1.546 & 62.62\%(-0.04\%) \\
			\multirow{2}{*}{$T=2$} & GPipe &1.549 &  63.24\% ($\sim$)\\
			& XPipe & \textbf{1.542} & \textbf{63.54\%}(+0.30\%) \\
			\multirow{2}{*}{$T=4$} &GPipe & 1.600 & 63.28\%($\sim$)   \\
			& XPipe & \textbf{1.596}& \textbf{63.72\%}(+0.44\%)\\
			%XPipe &\textbf{0.269} & \textbf{92.18\%} (+0.08\%)  \\
			\hline
			\hline
			\multirow{2}{*}{$T=1$} & GPipe & 1.508 & 63.24\% ($\sim$) \\
			& XPipe & \textbf{1.438} & \textbf{64.04\%}(+0.80\%) \\
			\multirow{2}{*}{$T=2$} & GPipe &1.495 & 64.60\% \\
			& XPipe & \textbf{1.459} & \textbf{65.06\%}(+0.46\%) \\
			\multirow{2}{*}{$T=4$} & GPipe &1.560 & 64.08\%   \\
			& XPipe &\textbf{1.540} & \textbf{64.82\%}(+0.74\%) \\
			%XPipe &\textbf{0.257} & 93.21\% (-0.05\%)  \\
			\hline
		\end{tabular}
		\label{tab:tinyimagenet}
	\end{center}
\end{table}

\noindent\textbf{Throughput Study} In this section, we compared the training speed of XPipe with that of PipeDream, SpecTrain, and GPipe using 2 and 4 GPUs, respectively. Here training speed is measured through throughput, which is defined by the number of training images per second. For PipeDream, SpecTrain, and XPipe, the throughput measurement refers to the number of the per-second training images throughout the \emph{steady phase}. We divided the comparison into two groups. For the first group, we compared the throughput of XPipe with that of PipeDream and SpecTrain. We selected VGG-16 and Inception-V3 as the benchmark network and trained them on the CIFAR-10 dataset for one epoch. The mini-batch size for all evaluated approaches was 128. In the first group, we always trained XPipe with $T=1$ to isolate the mini-batch partition effects. For the second group, we compared the throughput of XPipe with that of GPipe by considering different mini-batch partitions (\ie,  $T=1$, $T=2$, and $T=4$). We selected Inception-V3 and ResNet-101 as the benchmark network and trained them on the Tiny ImageNet for one epoch. The mini-batch sizes for both GPipe and XPipe on 2-GPU and 4-GPU systems were set to $50\times T$ and $100\times T$, respectively.

\begin{figure}[htbp]
	\centering
	\subfigure{\includegraphics[width=0.38\textwidth, height=4.2cm]{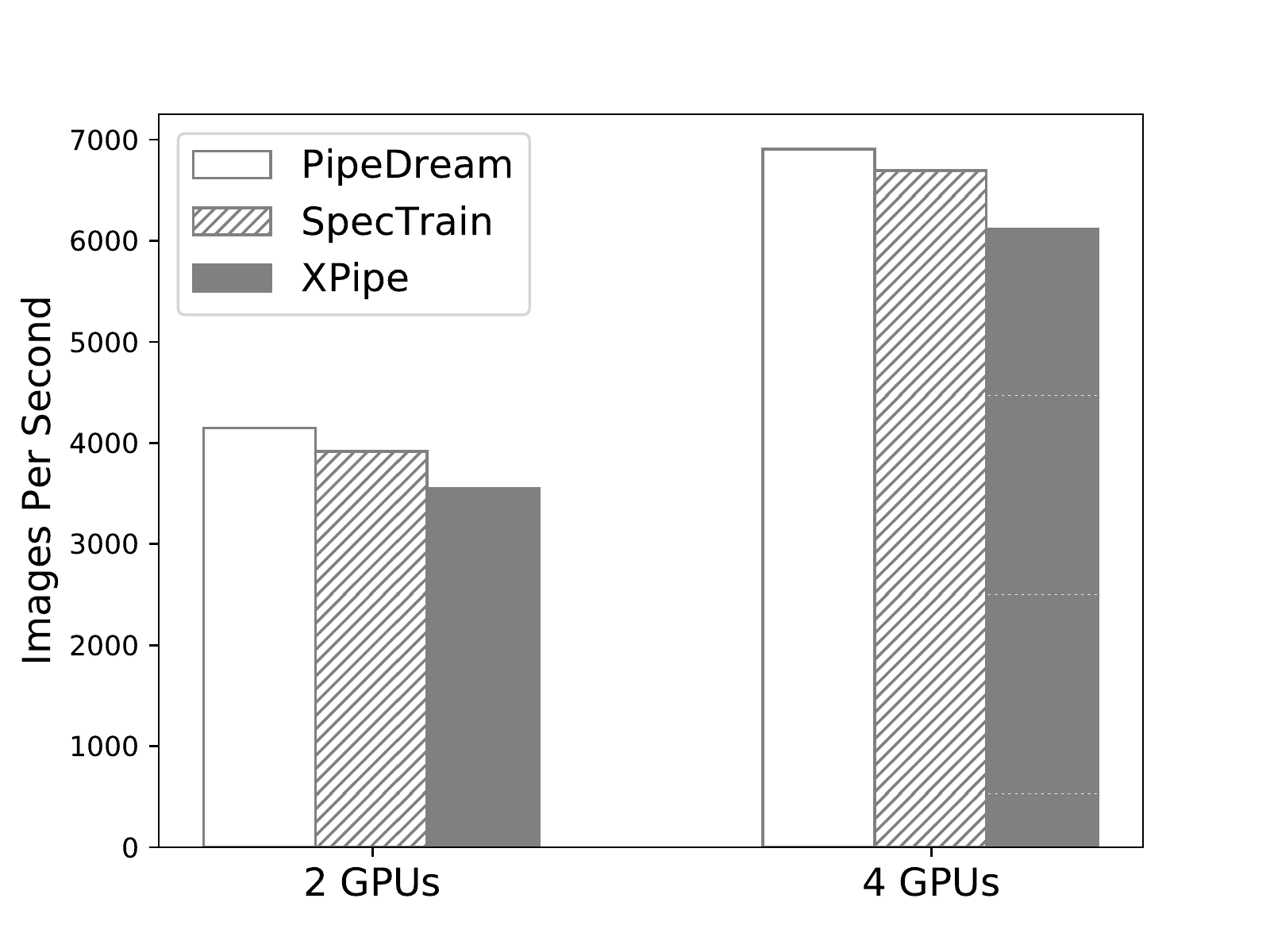}\label{vgg16_througput}}
	\subfigure{\includegraphics[width=0.38\textwidth, height=4.2cm]{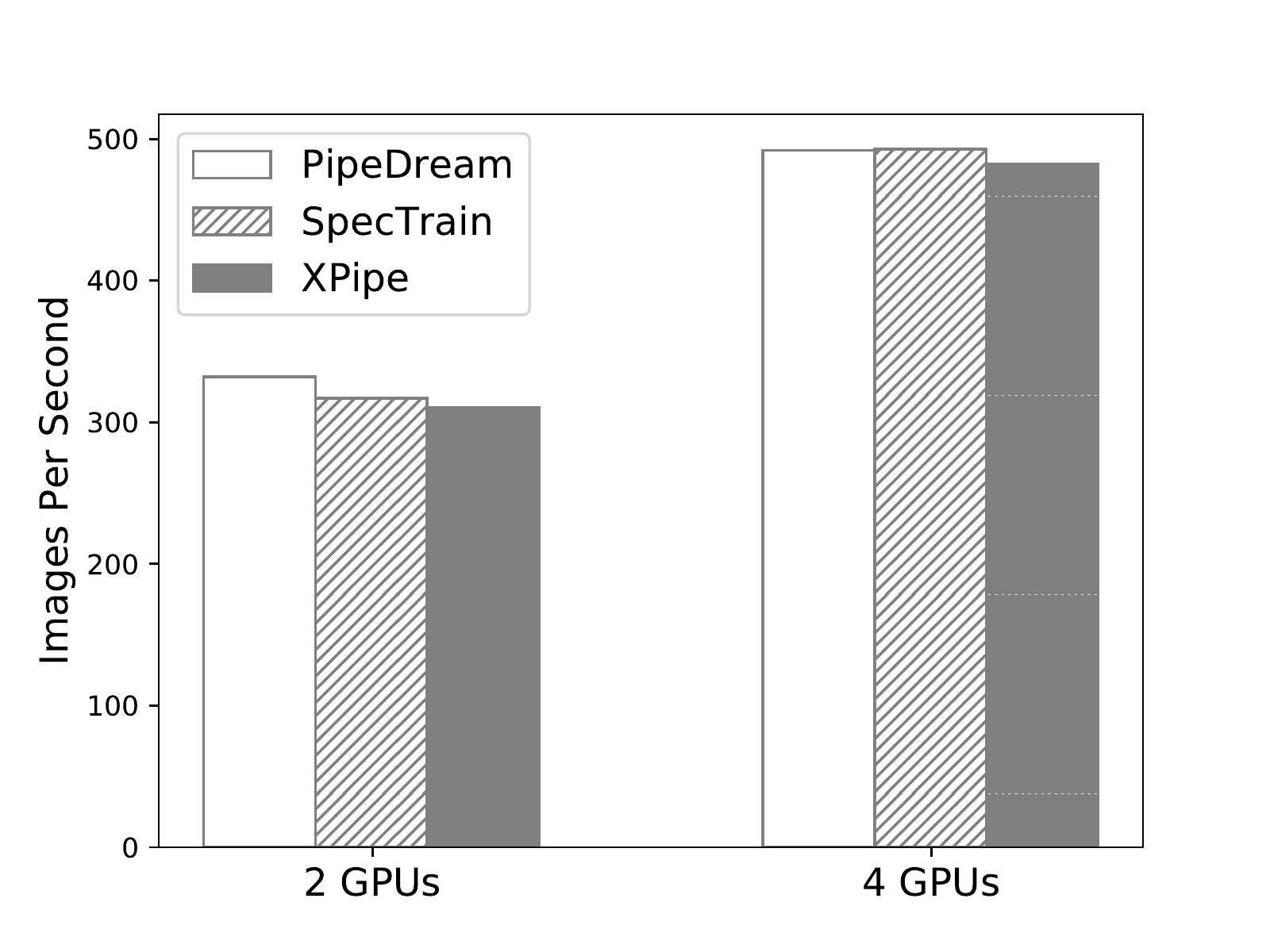}\label{inceptionv3_throughput}}
	%\subfigure[2-GPU system]{\includegraphics[width=0.34\textwidth, height=4.0cm]{figure/xpipe_gpipe_throughput_2gpu}\label{sync_2gpu_inceptionv3_througput}}
	%\subfigure[4-GPU system]{\includegraphics[width=0.34\textwidth, height=4.0cm]{figure/xpipe_gpipe_throughput_4gpu}\label{sync_4gpu_inceptionv3_throughput}}
	%\subfigure[2-GPU system]{\includegraphics[width=0.34\textwidth, height=4.0cm]{figure/xpipe_gpipe_resnet101_throughput_2gpu}\label{sync_resnet101_2gpu_througput}}
	%\subfigure[4-GPU system]{\includegraphics[width=0.34\textwidth, height=4.0cm]{figure/xpipe_gpipe_resnet101_throughput_4gpu}\label{sync_resnet101_4gpu_throughput}}
	%\subfigure[Validation Loss]{\includegraphics[width=0.24\textwidth, height=3.8cm]{figure/sgd_resnet101_4_loss}\label{sgd_resnet101_loss}}
	%\subfigure[Validation Top-1 Accuracy]{\includegraphics[width=0.24\textwidth, height=3.8cm]{figure/sgd_resnet101_4_top1}\label{sgd_resnet101_top1}}
	\caption{Throughputs of PipeDream, SpecTrain and XPipe on 2-GPU and 4-GPU systems. Top: throughput results for training VGG-16; bottom: throughput results for training Inception-V3.}
	\label{fig:asyc-throughput}
\end{figure}
\begin{figure}[htbp]
	\centering
	%\subfigure{\includegraphics[width=0.34\textwidth, height=4.0cm]{figure/vgg16_throughput}\label{vgg16_througput}}
	%\subfigure{\includegraphics[width=0.34\textwidth, height=4.0cm]{figure/inceptionv3_throughput}\label{inceptionv3_throughput}}
	\subfigure{\includegraphics[width=0.38\textwidth, height=4.2cm]{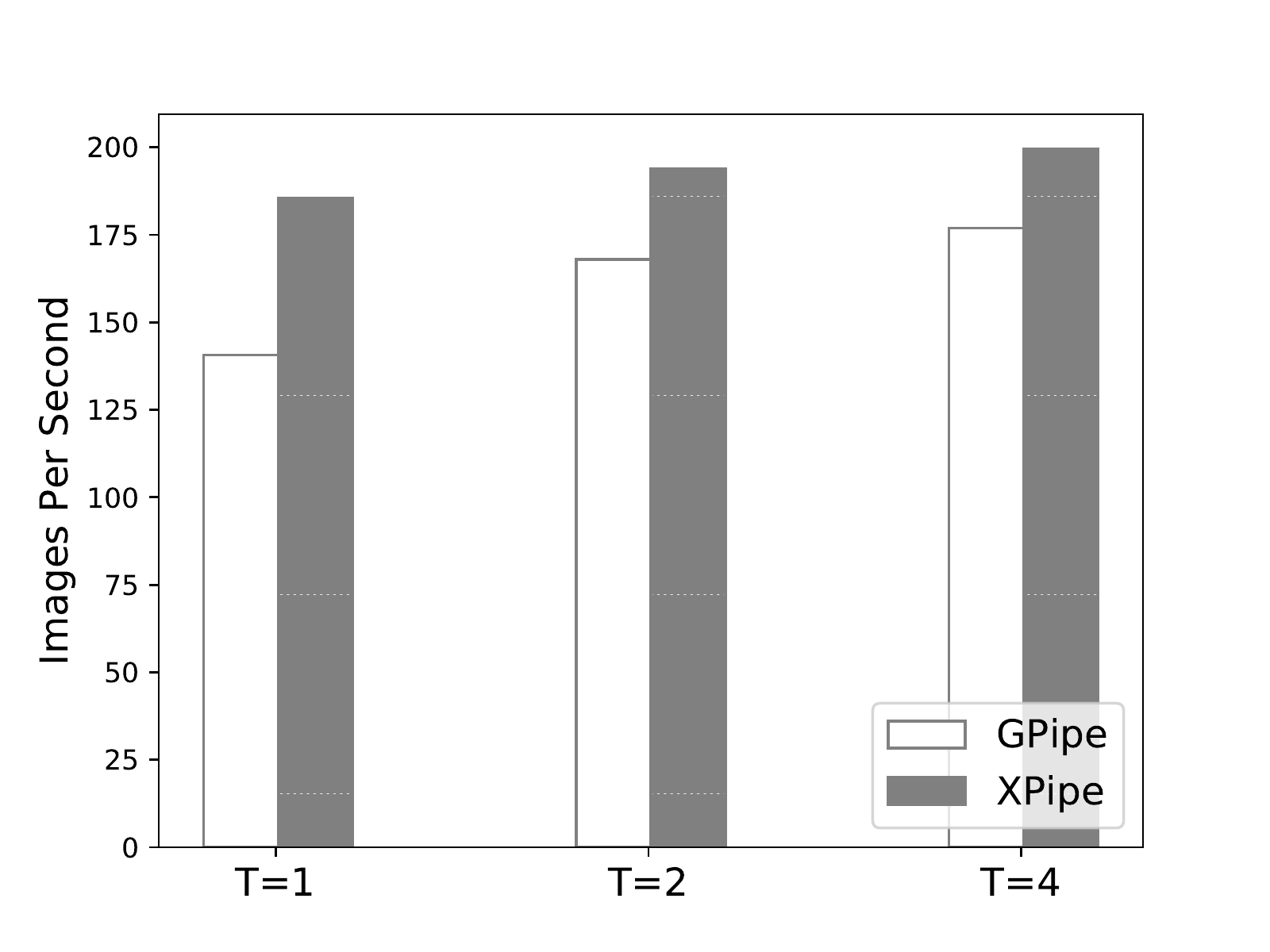}\label{sync_2gpu_inceptionv3_througput}}
	\subfigure{\includegraphics[width=0.38\textwidth, height=4.2cm]{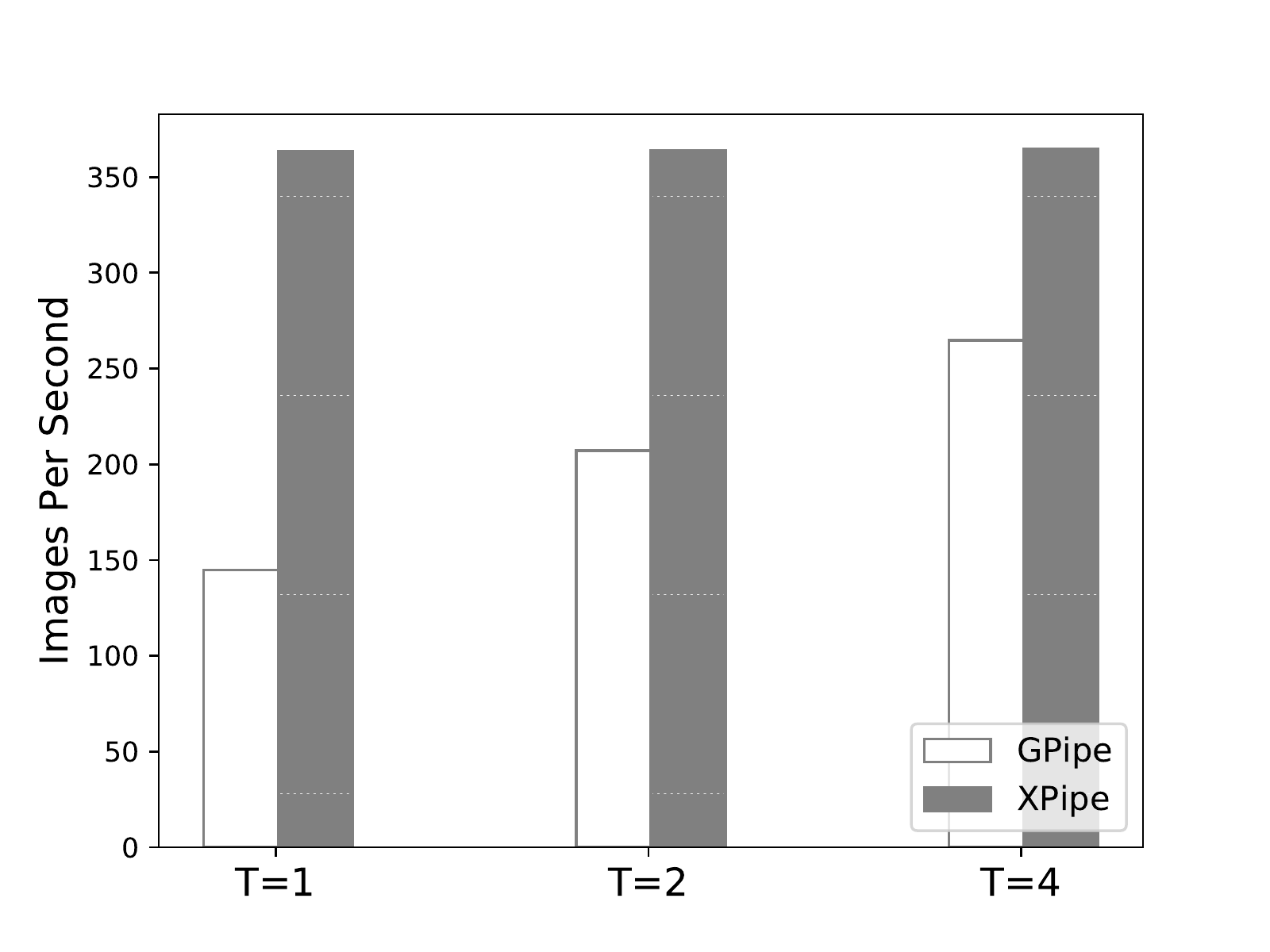}\label{sync_4gpu_inceptionv3_throughput}}
	%\subfigure[2-GPU system]{\includegraphics[width=0.34\textwidth, height=4.0cm]{figure/xpipe_gpipe_resnet101_throughput_2gpu}\label{sync_resnet101_2gpu_througput}}
	%\subfigure[4-GPU system]{\includegraphics[width=0.34\textwidth, height=4.0cm]{figure/xpipe_gpipe_resnet101_throughput_4gpu}\label{sync_resnet101_4gpu_throughput}}
	%\subfigure[Validation Loss]{\includegraphics[width=0.24\textwidth, height=3.8cm]{figure/sgd_resnet101_4_loss}\label{sgd_resnet101_loss}}
	%\subfigure[Validation Top-1 Accuracy]{\includegraphics[width=0.24\textwidth, height=3.8cm]{figure/sgd_resnet101_4_top1}\label{sgd_resnet101_top1}}
	\caption{Throughputs of GPipe and XPipe for training Inception-V3 with $T=1$, $T=2$, and $T=4$. Top: throughput results on 2-GPU system; bottom: throughput results on 4-GPU system.}
	\label{fig:syc-throughput-inceptionv3}
\end{figure}
\begin{figure}[htbp]
	\centering
	%\subfigure{\includegraphics[width=0.34\textwidth, height=4.0cm]{figure/vgg16_throughput}\label{vgg16_througput}}
	%\subfigure{\includegraphics[width=0.34\textwidth, height=4.0cm]{figure/inceptionv3_throughput}\label{inceptionv3_throughput}}
	%\subfigure[2-GPU system]{\includegraphics[width=0.34\textwidth, height=4.0cm]{figure/xpipe_gpipe_throughput_2gpu}\label{sync_2gpu_inceptionv3_througput}}
	%\subfigure[4-GPU system]{\includegraphics[width=0.34\textwidth, height=4.0cm]{figure/xpipe_gpipe_throughput_4gpu}\label{sync_4gpu_inceptionv3_throughput}}
	\subfigure{\includegraphics[width=0.38\textwidth, height=4.2cm]{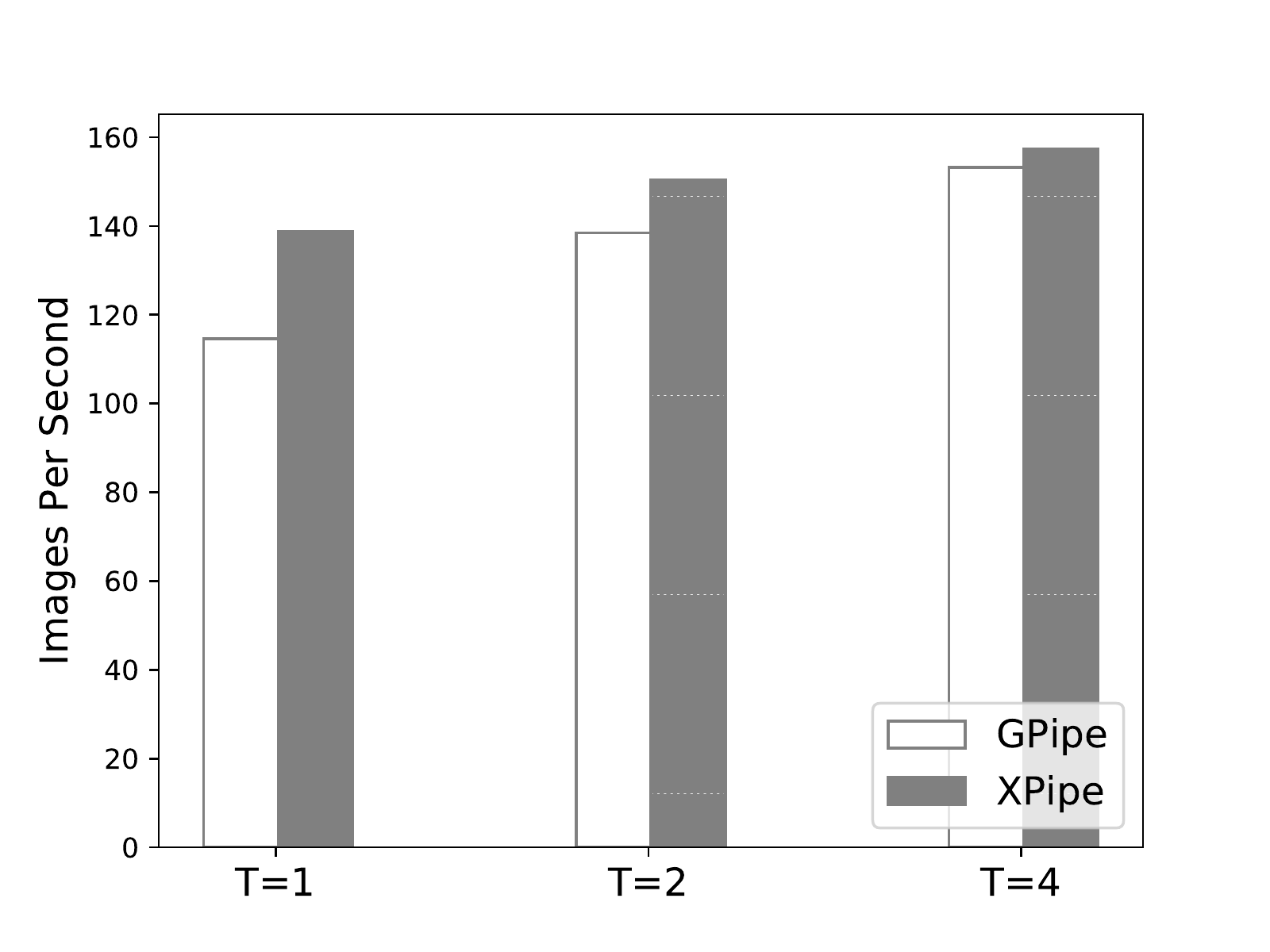}\label{sync_resnet101_2gpu_througput}}
	\subfigure{\includegraphics[width=0.38\textwidth, height=4.2cm]{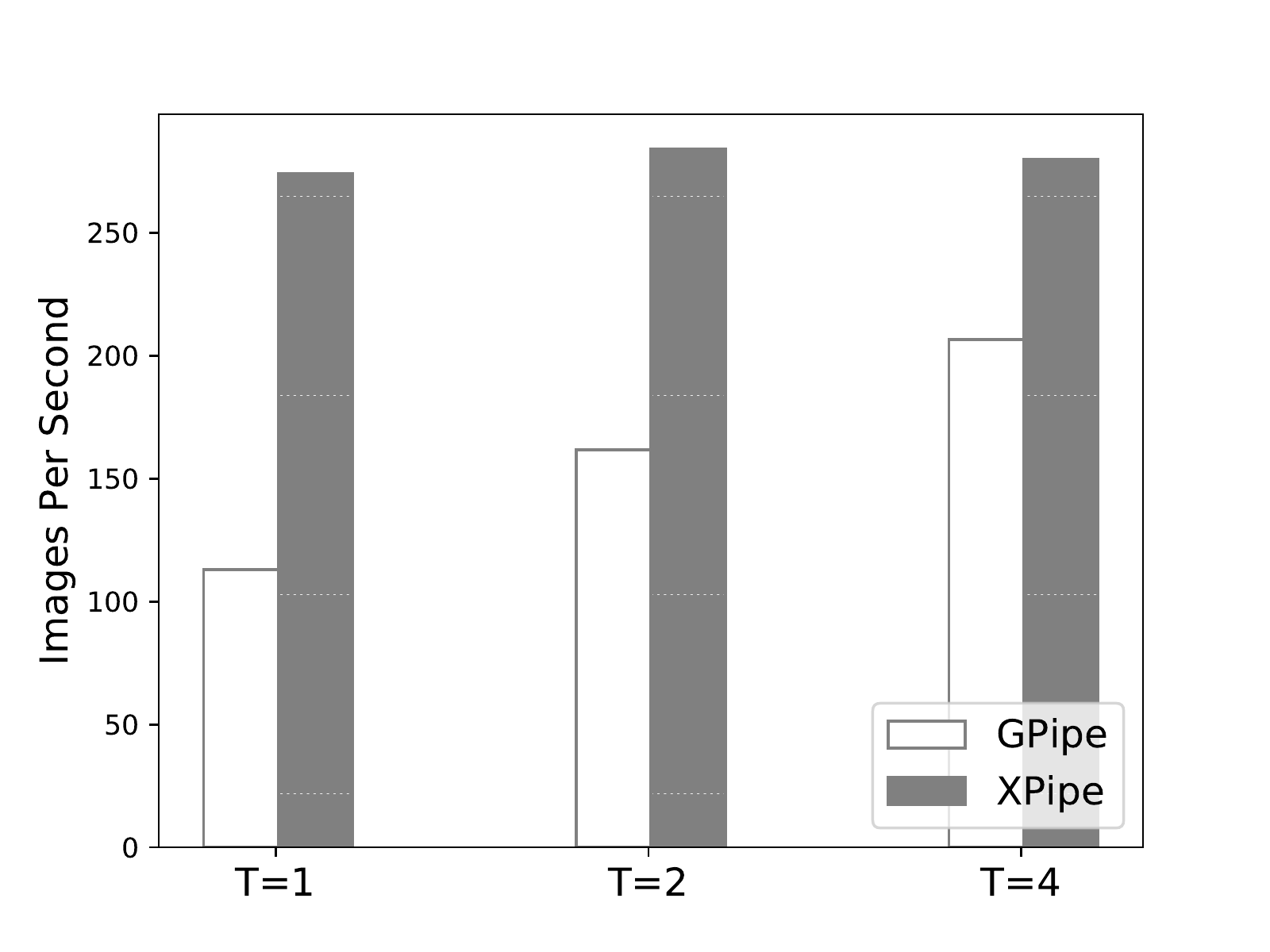}\label{sync_resnet101_4gpu_throughput}}
	%\subfigure[Validation Loss]{\includegraphics[width=0.24\textwidth, height=3.8cm]{figure/sgd_resnet101_4_loss}\label{sgd_resnet101_loss}}
	%\subfigure[Validation Top-1 Accuracy]{\includegraphics[width=0.24\textwidth, height=3.8cm]{figure/sgd_resnet101_4_top1}\label{sgd_resnet101_top1}}
	\caption{Throughputs of GPipe and XPipe for training ResNet-101 with $T=1$, $T=2$, and $T=4$. Top: throughput results on 2-GPU system; bottom: throughput results on 4-GPU system.}
	\label{fig:syc-throughput-resnet101}
\end{figure}

Figure \ref{fig:asyc-throughput} illustrates the throughput results of the first group; Figures \ref{fig:syc-throughput-inceptionv3} and \ref{fig:syc-throughput-resnet101} show the throughput results of the second group. It is worth noting that the experiments were conducted to compare the throughput of XPipe with that of other pipeline approaches under the same hyper-parameter setting. All the evaluated pipeline approaches can consistently obtain higher throughput when better model partition approach or elaborated hyper-parameter tuning (e.g., mini-batch size or $T$) is applied.

We can reach to following conclusions based on the observation of the throughput results. First, the throughput results in Figure \ref{fig:asyc-throughput} show that the throughput of XPipe is slightly inferior to that of PipeDream and SpecTrain despite all of them adopt the same pipeline structure. This is because XPipe takes advantage of a more computation-intensive weight prediction strategy to guarantee effective learning. %Second, as with GPipe, XPipe scales up the mini-batch size easily. It is very reasonable because both GPipe and XPipe use the fine-grained micro-batch as the basic data processing unit in the pipeline training.
Second, Figures \ref{fig:syc-throughput-inceptionv3} and \ref{fig:syc-throughput-resnet101} show that the throughput of GPipe is very sensitive to the choice of $T$ especially when training on the 4-GPU system. This is because the pipeline structure of GPipe varies with the setting of $T$, and different settings of $T$ give rise to different proportions of idle time. Contrastly, the pipeline structure of XPipe is independent of the setting of $T$. XPipe can always make all GPUs concurrently train the DNN model after the \emph{steady phase} starts. Therefore, XPipe trains very fast and can consistently achieve high throughput, regardless of the setting of $T$. For Inception-V3, XPipe provides an average of 20.0\% (up to 31.9\%) and 88.1\% (up to 150.8\%) throughput improvement over GPipe on 2-GPU and 4-GPU systems, respectively. For ResNet-101, XPipe provides an average of 10.8\% (up to 21.2\%) and 84.6\% (up to 142.7\%) throughput improvement over GPipe on 2-GPU and 4-GPU systems, respectively.

%XPipe always trains  faster than XPipe, in dependence of the selection of $T$.

%Second, PipeDream, SpecTrain and XPipe are able to obtain much higher throughput than GPipe, which demonstrates that the asynchronous pipeline training produces higher GPU utilization than the synchronous pipeline training. In special, Third,

\noindent\textbf{Robustness Study} In this section, we study the robustness of XPipe by using another two popular optimization methods for pipeline training: RMSProp~\cite{tieleman2012lecture} and Adam~\cite{kingma2014adam}. We again trained GPipe with $T=1$ to simulate the behavior of the naive model parallel approach. We regarded the results of it as the baseline. We selected VGG-16 as the benchmark network and trained it on CIFAR-10 for 50 epochs using 4 GPUs. We fixed the learning rate as 1e-4. The mini-batch size for all approaches was 128. For RMSProp, we set the momentum value to 0.9; for Adam, we set the exponential decay rates for the first and second momentum estimates to 0.9 and 0.999, respectively.

\begin{figure}[htbp]
	\centering
	%\subfigure[Momentum SGD]{\includegraphics[width=0.30\textwidth, height=3.8cm]{figure/sgd_vgg16_4_loss}\label{sgd_vgg16_loss}}
	%\subfigure[RMSProp]{\includegraphics[width=0.30\textwidth, height=3.8cm]{figure/rmsprop_vgg16_4_loss}\label{rmsprop_vgg16_loss}}
	%\subfigure[Adam]{\includegraphics[width=0.20\textwidth, height=3.6cm]{figure/adam_vgg16_4_loss}\label{adam_vgg16_loss}}	
	%\subfigure[Momentum SGD]{\includegraphics[width=0.20\textwidth, height=3.6cm]{figure/sgd_vgg16_4_top1}\label{sgd_vgg16_top1}}
	\subfigure{\includegraphics[width=0.23\textwidth, height=3.8cm]{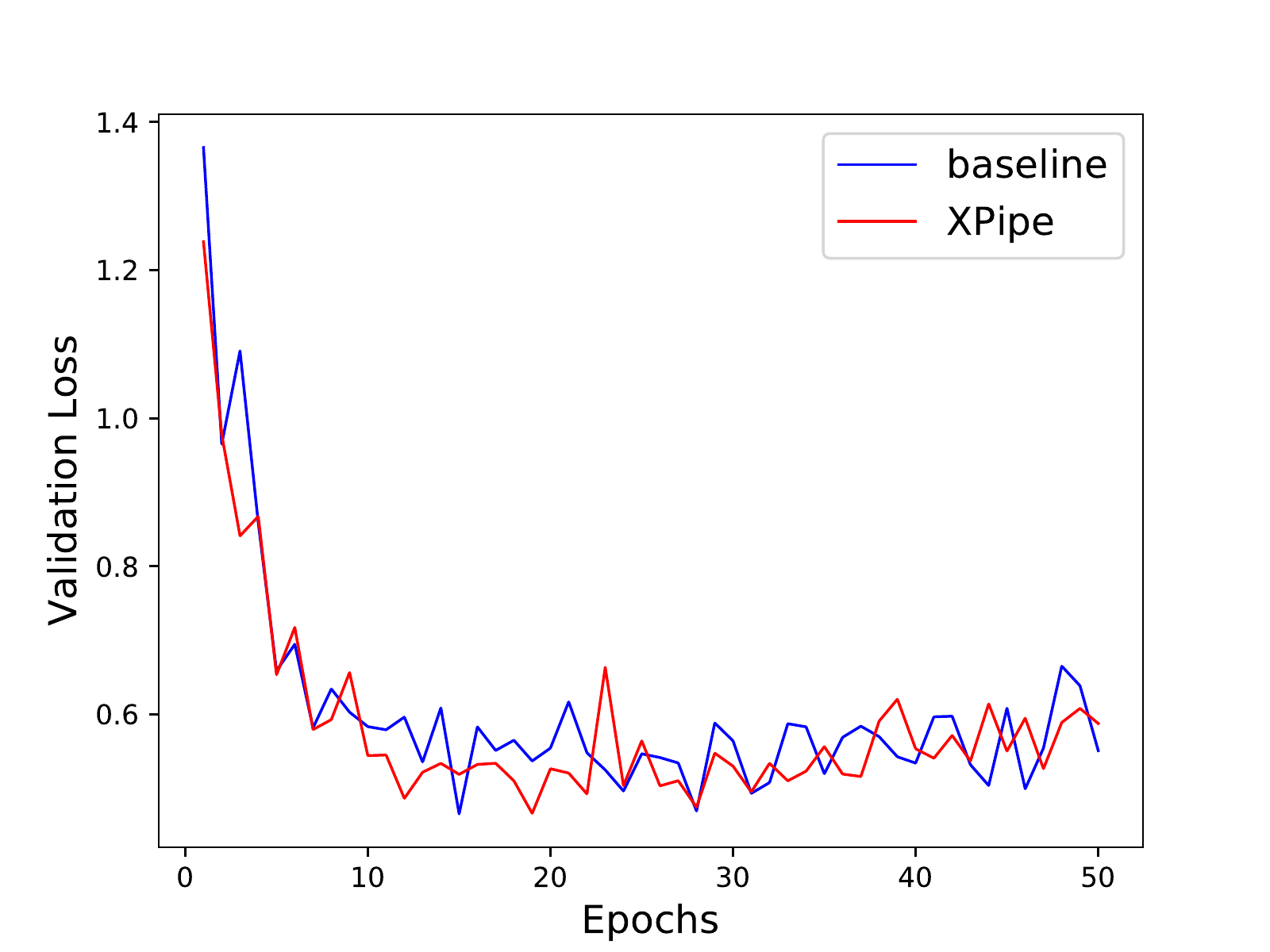}\label{rmsprop_vgg16_loss}}	
	\subfigure{\includegraphics[width=0.23\textwidth, height=3.8cm]{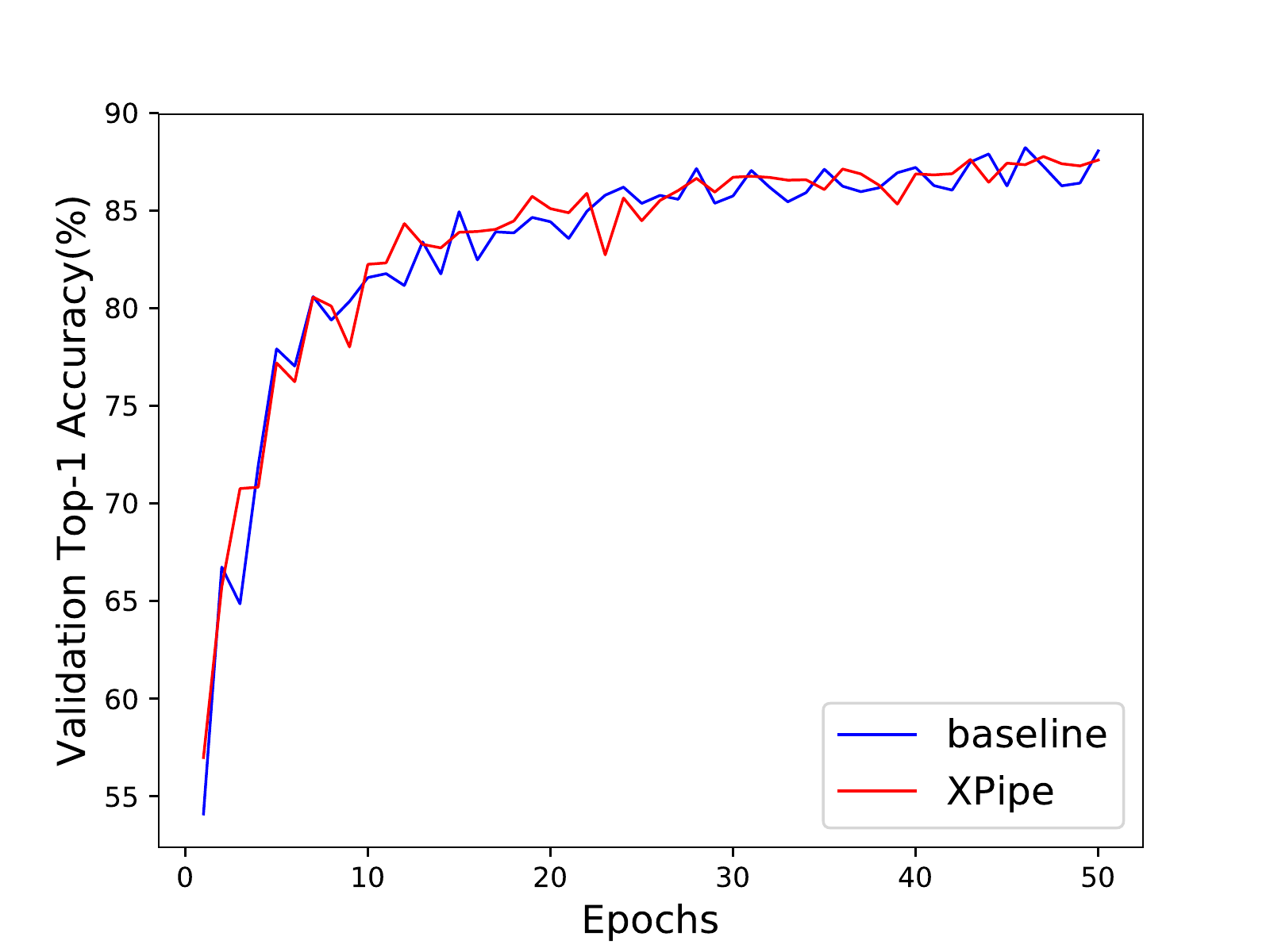}\label{rmsprop_vgg16_top1}}
	\caption{Learning curves when using RMSProp optimizer. }
	\label{fig:robustness-rmsprop}
\end{figure}

\begin{figure}[htbp]
	\centering
	%\subfigure[Momentum SGD]{\includegraphics[width=0.30\textwidth, height=3.8cm]{figure/sgd_vgg16_4_loss}\label{sgd_vgg16_loss}}
	%\subfigure[RMSProp]{\includegraphics[width=0.30\textwidth, height=3.8cm]{figure/rmsprop_vgg16_4_loss}\label{rmsprop_vgg16_loss}}
	%\subfigure[Adam]{\includegraphics[width=0.20\textwidth, height=3.6cm]{figure/adam_vgg16_4_loss}\label{adam_vgg16_loss}}	
	%\subfigure[Momentum SGD]{\includegraphics[width=0.20\textwidth, height=3.6cm]{figure/sgd_vgg16_4_top1}\label{sgd_vgg16_top1}}
	\subfigure{\includegraphics[width=0.23\textwidth, height=3.8cm]{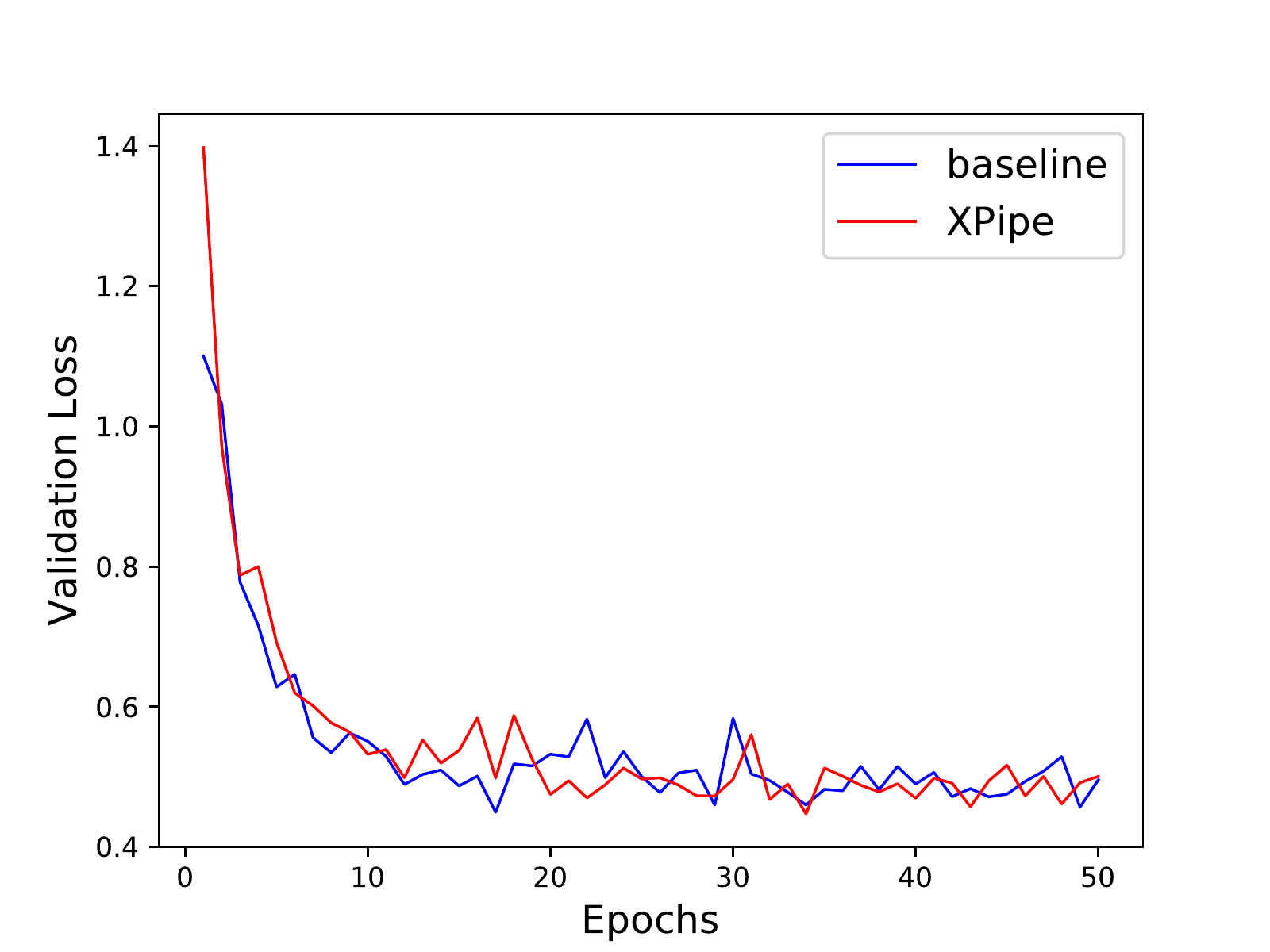}\label{adam_vgg16_loss}}	
	\subfigure{\includegraphics[width=0.23\textwidth, height=3.8cm]{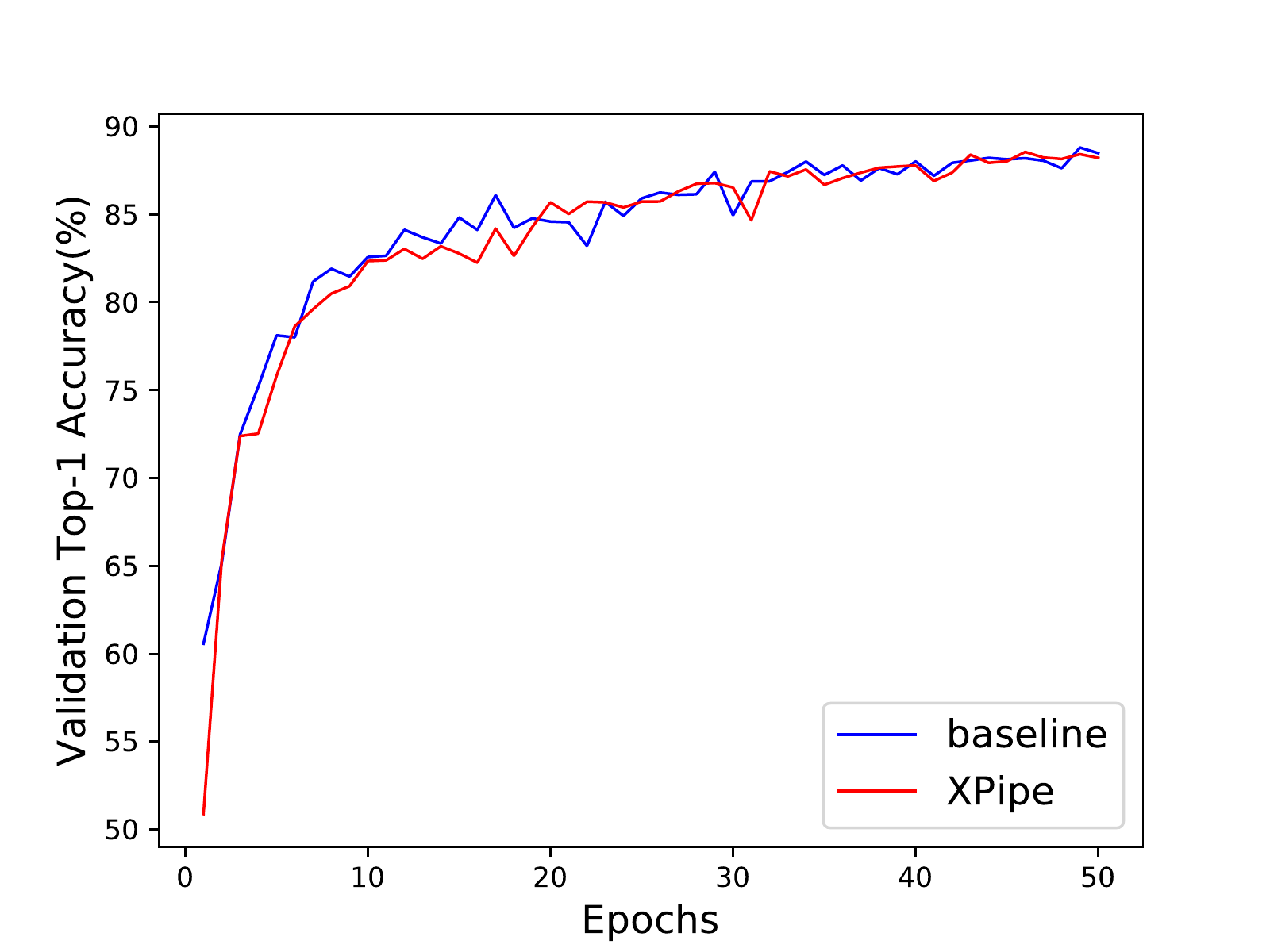}\label{adam_vgg16_top1}}
	\caption{Learning curves when using Adam optimizer.}
	\label{fig:robustness-adam}
\end{figure}

Figures \ref{fig:robustness-rmsprop} and \ref{fig:robustness-adam} show the experimental results when using RMSProp and Adam as the optimization method, respectively. In both Figures \ref{fig:robustness-rmsprop} and \ref{fig:robustness-adam}, the left figure depicts the validation loss over epochs and the right figure shows the validation accuracy (top-1, in \%) over epochs. The experimental results demonstrate the effectiveness of XPipe. For using either RMSProp or Adam as the optimization method, the learning curves of XPipe converge quickly and match well with that of the baseline. This demonstrates the robustness of Adam-based weight prediction strategy. XPipe can always guarantee learning-effective training, and the results are independent of the optimizer method used.
%At the same time, XPipe provides very comparable (sometimes better) top-1 validation accuracy as GPipe regardless of the optimization method used. In contrast, PipeDream and SpecTrain suffers from top-1 accuracy drop for any optimization method. On average, in comparison with the baseline, PipeDream and SpecTrain get 3.17\% (up to 6.92\%) and 0.45\% (up to 0.80\%) accuracy loss, respectively.

\section{Conclusions}
%GPU has been seen as a primary tool to speed up DNN training. When using multiple GPUs to train a DNN in parallel, load balance across GPUs becomes a key factor to achieve high GPU utilization.
In this work, we propose an efficient asynchronous pipeline model parallelism method called XPipe. XPipe interweaves the pipeline training of micro-batches belonging to different mini-batches to ensure that each GPU concurrently and continuously trains the DNN model, thereby providing pretty high throughput. Most importantly, the novel weight prediction scheme makes XPipe effectively address the inconsistency and staleness issues in the asynchronous pipeline training. Overall, XPipe provides pretty high throughput, scales up mini-batch size easily, and achieves very comparable accuracy (even slightly better) as its state-of-the-art synchronous counterpart.

\section*{Acknowledgments}

The work was partially supported by the China Scholarship Council (CSC) and Major State Research Development Program, China (2016YFB0201305). Lei Guan thanks Zhihui Yang, Tao Sun, and Bao Wang for stimulating discussions.

\bibliographystyle{IEEEtran}
\bibliography{IEEEabrv, mybibfile}

\begin{thebibliography}{10}
\providecommand{\url}[1]{#1}
\csname url@samestyle\endcsname
\providecommand{\newblock}{\relax}
\providecommand{\bibinfo}[2]{#2}
\providecommand{\BIBentrySTDinterwordspacing}{\spaceskip=0pt\relax}
\providecommand{\BIBentryALTinterwordstretchfactor}{4}
\providecommand{\BIBentryALTinterwordspacing}{\spaceskip=\fontdimen2\font plus
\BIBentryALTinterwordstretchfactor\fontdimen3\font minus
  \fontdimen4\font\relax}
\providecommand{\BIBforeignlanguage}[2]{{%
\expandafter\ifx\csname l@#1\endcsname\relax
\typeout{** WARNING: IEEEtran.bst: No hyphenation pattern has been}%
\typeout{** loaded for the language `#1'. Using the pattern for}%
\typeout{** the default language instead.}%
\else
\language=\csname l@#1\endcsname
\fi
#2}}
\providecommand{\BIBdecl}{\relax}
\BIBdecl

\bibitem{szegedy2017inception}
C.~Szegedy, S.~Ioffe, V.~Vanhoucke, and A.~A. Alemi, ``Inception-v4,
  inception-resnet and the impact of residual connections on learning,'' in
  \emph{Thirty-First AAAI Conference on Artificial Intelligence}, 2017.

\bibitem{karpathy2014large}
A.~Karpathy, G.~Toderici, S.~Shetty, T.~Leung, R.~Sukthankar, and L.~Fei-Fei,
  ``Large-scale video classification with convolutional neural networks,'' in
  \emph{Proceedings of the IEEE conference on Computer Vision and Pattern
  Recognition}, 2014, pp. 1725--1732.

\bibitem{kalchbrenner2014convolutional}
N.~Kalchbrenner, E.~Grefenstette, and P.~Blunsom, ``A convolutional neural
  network for modelling sentences,'' \emph{arXiv preprint arXiv:1404.2188},
  2014.

\bibitem{wu2016google}
Y.~Wu, M.~Schuster, Z.~Chen, Q.~V. Le, M.~Norouzi, W.~Macherey, M.~Krikun,
  Y.~Cao, Q.~Gao, K.~Macherey \emph{et~al.}, ``Google's neural machine
  translation system: Bridging the gap between human and machine translation,''
  \emph{arXiv preprint arXiv:1609.08144}, 2016.

\bibitem{afouras2018deep}
T.~Afouras, J.~S. Chung, A.~Senior, O.~Vinyals, and A.~Zisserman, ``Deep
  audio-visual speech recognition,'' \emph{IEEE transactions on pattern
  analysis and machine intelligence}, 2018.

\bibitem{lee2009unsupervised}
H.~Lee, P.~Pham, Y.~Largman, and A.~Y. Ng, ``Unsupervised feature learning for
  audio classification using convolutional deep belief networks,'' in
  \emph{Advances in neural information processing systems}, 2009, pp.
  1096--1104.

\bibitem{you2018imagenet}
Y.~You, Z.~Zhang, C.-J. Hsieh, J.~Demmel, and K.~Keutzer, ``Imagenet training
  in minutes,'' in \emph{Proceedings of the 47th International Conference on
  Parallel Processing}.\hskip 1em plus 0.5em minus 0.4em\relax ACM, 2018, p.~1.

\bibitem{shazeer2017outrageously}
N.~Shazeer, A.~Mirhoseini, K.~Maziarz, A.~Davis, Q.~Le, G.~Hinton, and J.~Dean,
  ``Outrageously large neural networks: The sparsely-gated mixture-of-experts
  layer,'' \emph{arXiv preprint arXiv:1701.06538}, 2017.

\bibitem{deng2009imagenet}
J.~Deng, W.~Dong, R.~Socher, L.-J. Li, K.~Li, and L.~Fei-Fei, ``Imagenet: A
  large-scale hierarchical image database,'' in \emph{2009 IEEE conference on
  computer vision and pattern recognition}.\hskip 1em plus 0.5em minus
  0.4em\relax Ieee, 2009, pp. 248--255.

\bibitem{hinton2015distilling}
G.~Hinton, O.~Vinyals, and J.~Dean, ``Distilling the knowledge in a neural
  network,'' \emph{arXiv preprint arXiv:1503.02531}, 2015.

\bibitem{krasin2017openimages}
I.~Krasin, T.~Duerig, N.~Alldrin, V.~Ferrari, S.~Abu-El-Haija, A.~Kuznetsova,
  H.~Rom, J.~Uijlings, S.~Popov, A.~Veit \emph{et~al.}, ``Openimages: A public
  dataset for large-scale multi-label and multi-class image classification,''
  \emph{Dataset available from https://github. com/openimages}, vol.~2, p.~3,
  2017.

\bibitem{taigman2014deepface}
Y.~Taigman, M.~Yang, M.~Ranzato, and L.~Wolf, ``Deepface: Closing the gap to
  human-level performance in face verification,'' in \emph{Proceedings of the
  IEEE conference on computer vision and pattern recognition}, 2014, pp.
  1701--1708.

\bibitem{ben2018demystifying}
T.~Ben-Nun and T.~Hoefler, ``Demystifying parallel and distributed deep
  learning: An in-depth concurrency analysis,'' \emph{arXiv preprint
  arXiv:1802.09941}, 2018.

\bibitem{lee2014model}
S.~Lee, J.~K. Kim, X.~Zheng, Q.~Ho, G.~A. Gibson, and E.~P. Xing, ``On model
  parallelization and scheduling strategies for distributed machine learning,''
  in \emph{Advances in neural information processing systems}, 2014, pp.
  2834--2842.

\bibitem{huang2019gpipe}
Y.~Huang, Y.~Cheng, A.~Bapna, O.~Firat, D.~Chen, M.~Chen, H.~Lee, J.~Ngiam,
  Q.~V. Le, Y.~Wu \emph{et~al.}, ``Gpipe: Efficient training of giant neural
  networks using pipeline parallelism,'' in \emph{Advances in Neural
  Information Processing Systems}, 2019, pp. 103--112.

\bibitem{narayanan2019pipedream}
D.~Narayanan, A.~Harlap, A.~Phanishayee, V.~Seshadri, N.~R. Devanur, G.~R.
  Ganger, P.~B. Gibbons, and M.~Zaharia, ``Pipedream: generalized pipeline
  parallelism for dnn training,'' in \emph{Proceedings of the 27th ACM
  Symposium on Operating Systems Principles}.\hskip 1em plus 0.5em minus
  0.4em\relax ACM, 2019, pp. 1--15.

\bibitem{chen2018efficient}
C.-C. Chen, C.-L. Yang, and H.-Y. Cheng, ``Efficient and robust parallel dnn
  training through model parallelism on multi-gpu platform,'' \emph{arXiv
  preprint arXiv:1809.02839}, 2018.

\bibitem{dean2012large}
J.~Dean, G.~Corrado, R.~Monga, K.~Chen, M.~Devin, M.~Mao, M.~Ranzato,
  A.~Senior, P.~Tucker, K.~Yang \emph{et~al.}, ``Large scale distributed deep
  networks,'' in \emph{Advances in neural information processing systems},
  2012, pp. 1223--1231.

\bibitem{huo2018training}
Z.~Huo, B.~Gu, and H.~Huang, ``Training neural networks using features
  replay,'' in \emph{Advances in Neural Information Processing Systems}, 2018,
  pp. 6659--6668.

\bibitem{huo2018decoupled}
Z.~Huo, B.~Gu, Q.~Yang, and H.~Huang, ``Decoupled parallel backpropagation with
  convergence guarantee,'' \emph{arXiv preprint arXiv:1804.10574}, 2018.

\bibitem{pal2019optimizing}
S.~Pal, E.~Ebrahimi, A.~Zulfiqar, Y.~Fu, V.~Zhang, S.~Migacz, D.~Nellans, and
  P.~Gupta, ``Optimizing multi-gpu parallelization strategies for deep learning
  training,'' \emph{IEEE Micro}, vol.~39, no.~5, pp. 91--101, 2019.

\bibitem{petrowski1993performance}
A.~Petrowski, G.~Dreyfus, and C.~Girault, ``Performance analysis of a pipelined
  backpropagation parallel algorithm,'' \emph{IEEE Transactions on Neural
  Networks}, vol.~4, no.~6, pp. 970--981, 1993.

\bibitem{kamruzzaman2013load}
M.~Kamruzzaman, S.~Swanson, and D.~M. Tullsen, ``Load-balanced pipeline
  parallelism,'' in \emph{SC'13: Proceedings of the International Conference on
  High Performance Computing, Networking, Storage and Analysis}.\hskip 1em plus
  0.5em minus 0.4em\relax IEEE, 2013, pp. 1--12.

\bibitem{chen2012pipelined}
X.~Chen, A.~Eversole, G.~Li, D.~Yu, and F.~Seide, ``Pipelined back-propagation
  for context-dependent deep neural networks,'' in \emph{Thirteenth Annual
  Conference of the International Speech Communication Association}, 2012.

\bibitem{pittman2018exploring}
R.~Pittman, H.~Guan, X.~Shen, S.-H. Lim, and R.~M. Patton, ``Exploring flexible
  communications for streamlining dnn ensemble training pipelines,'' in
  \emph{Proceedings of the International Conference for High Performance
  Computing, Networking, Storage, and Analysis}.\hskip 1em plus 0.5em minus
  0.4em\relax IEEE Press, 2018, p.~64.

\bibitem{gaunt2017ampnet}
A.~L. Gaunt, M.~A. Johnson, M.~Riechert, D.~Tarlow, R.~Tomioka, D.~Vytiniotis,
  and S.~Webster, ``Ampnet: Asynchronous model-parallel training for dynamic
  neural networks,'' \emph{arXiv preprint arXiv:1705.09786}, 2017.

\bibitem{qian1999momentum}
N.~Qian, ``On the momentum term in gradient descent learning algorithms,''
  \emph{Neural networks}, vol.~12, no.~1, pp. 145--151, 1999.

\bibitem{kingma2014adam}
D.~P. Kingma and J.~Ba, ``Adam: A method for stochastic optimization,''
  \emph{arXiv preprint arXiv:1412.6980}, 2014.

\bibitem{paszke2017automatic}
A.~Paszke, S.~Gross, S.~Chintala, G.~Chanan, E.~Yang, Z.~DeVito, Z.~Lin,
  A.~Desmaison, L.~Antiga, and A.~Lerer, ``Automatic differentiation in
  pytorch,'' 2017.

\bibitem{mirhoseini2017device}
A.~Mirhoseini, H.~Pham, Q.~V. Le, B.~Steiner, R.~Larsen, Y.~Zhou, N.~Kumar,
  M.~Norouzi, S.~Bengio, and J.~Dean, ``Device placement optimization with
  reinforcement learning,'' in \emph{Proceedings of the 34th International
  Conference on Machine Learning-Volume 70}.\hskip 1em plus 0.5em minus
  0.4em\relax JMLR. org, 2017, pp. 2430--2439.

\bibitem{simonyan2014very}
K.~Simonyan and A.~Zisserman, ``Very deep convolutional networks for
  large-scale image recognition,'' \emph{arXiv preprint arXiv:1409.1556}, 2014.

\bibitem{he2016deep}
K.~He, X.~Zhang, S.~Ren, and J.~Sun, ``Deep residual learning for image
  recognition,'' in \emph{Proceedings of the IEEE conference on computer vision
  and pattern recognition}, 2016, pp. 770--778.

\bibitem{Szegedy_2016_CVPR}
C.~Szegedy, V.~Vanhoucke, S.~Ioffe, J.~Shlens, and Z.~Wojna, ``Rethinking the
  inception architecture for computer vision,'' in \emph{The IEEE Conference on
  Computer Vision and Pattern Recognition (CVPR)}, June 2016.

\bibitem{krizhevsky2009learning}
A.~Krizhevsky and G.~Hinton, ``Learning multiple layers of features from tiny
  images,'' Citeseer, Tech. Rep., 2009.

\bibitem{yao2015tiny}
L.~Yao and J.~Miller, ``Tiny imagenet classification with convolutional neural
  networks,'' \emph{CS 231N}, vol.~2, no.~5, p.~8, 2015.

\bibitem{tieleman2012lecture}
T.~Tieleman and G.~Hinton, ``Lecture 6.5-rmsprop: Divide the gradient by a
  running average of its recent magnitude,'' \emph{COURSERA: Neural networks
  for machine learning}, vol.~4, no.~2, pp. 26--31, 2012.

\end{thebibliography}

%\begin{thebibliography}{00}
%\bibitem{b1} G. Eason, B. Noble, and I. N. Sneddon, ``On certain integrals of Lipschitz-Hankel type involving products of Bessel functions,'' Phil. %Trans. Roy. Soc. London, vol. A247, pp. 529--551, April 1955.
%\bibitem{b2} J. Clerk Maxwell, A Treatise on Electricity and Magnetism, 3rd ed., vol. 2. Oxford: Clarendon, 1892, pp.68--73.
%\bibitem{b3} I. S. Jacobs and C. P. Bean, ``Fine particles, thin films and exchange anisotropy,'' in Magnetism, vol. III, G. T. Rado and H. Suhl, Eds. New York: Academic, 1963, pp. 271--350.
%\bibitem{b4} K. Elissa, ``Title of paper if known,'' unpublished.
%\bibitem{b5} R. Nicole, ``Title of paper with only first word capitalized,'' J. Name Stand. Abbrev., in press.
%\bibitem{b6} Y. Yorozu, M. Hirano, K. Oka, and Y. Tagawa, ``Electron spectroscopy studies on magneto-optical media and plastic substrate interface,'' IEEE Transl. J. Magn. Japan, vol. 2, pp. 740--741, August 1987 [Digests 9th Annual Conf. Magnetics Japan, p. 301, 1982].
%\bibitem{b7} M. Young, The Technical Writer's Handbook. Mill Valley, CA: University Science, 1989.
%\end{thebibliography}

\end{document}